**Ain't Nothing But a Survey? Using Large Language Models for Coding German Open-Ended Survey Responses on Survey Motivation**

*Leah von der Heyde,\* [1, 2] Anna-Carolina Haensch,[1, 2, 3] Bernd Weiß [4], Jessica Daikeler [4]*

[1] Social Data Science & AI Lab, LMU Munich
[2] Munich Center for Machine Learning
[3] University of Maryland, College Park
[4] GESIS – Leibniz Institute for the Social Sciences

*\* Corresponding author: l.heyde[at]lmu[dot]de*

**Abstract**

The recent development and wider accessibility of large language models (LLMs) have spurred discussions about how these language models can be used in survey research, including classifying open-ended survey responses. Due to their linguistic capacities, it is possible that LLMs are an efficient alternative to time-consuming manual coding and the pre-training of supervised machine learning models. As most existing research on this topic has focused on English-language responses relating to non-complex topics or on single LLMs, it is unclear whether its findings generalize and how the quality of these classifications compares to established methods. In this study, we investigate to what extent different LLMs can be used to code open-ended survey responses in other contexts, using German data on reasons for survey participation as an example. We compare several state-of-the-art LLMs of the GPT, Llama, and Mistral families, and several prompting approaches, including zero- and few-shot prompting and fine-tuning, and evaluate the LLMs' performance by using human expert codings. Overall performance differs greatly between LLMs, and only a fine-tuned LLM achieves satisfactory levels of predictive performance. Performance differences between prompting approaches are conditional on the LLM used. Finally, LLMs' unequal classification performance across different categories of reasons for survey participation results in different categorical distributions when not using fine-tuning. We discuss the implications of these findings, both for methodological research on coding open-ended responses and for their substantive analysis, and for practitioners processing or substantively analyzing such data. Finally, we highlight the many trade-offs researchers need to consider when choosing automated methods for open-ended response classification in the age of LLMs. In doing so, our study contributes to the growing body of research about the conditions under which LLMs can be efficiently, accurately, and reliably leveraged in survey research and their impact on data quality.

1. Introduction

The recent development and wider accessibility of large language models (LLMs) have spurred discussions about how these language models can be used in survey research. Potential applications span the entire survey lifecycle, including using LLMs for questionnaire design and pretesting (e.g., Götz et al., 2023), conducting interviews (e.g., Cuevas et al., 2023), synthesizing or imputing respondent data (e.g., Argyle et al., 2023; Kim & Lee, 2023), or detecting non-human respondents in online surveys (e.g., Lebrun et al., 2024). Due to their linguistic capacities, including their adaptability to different topics, the detection of nuance, implicitness, and intent in low-information multilingual textual input, and flexibility in generating textual output, LLMs also offer promising potential for classifying open-ended survey responses, which often are short and do not provide explicit context. For example, using LLMs for coding free-text social media data has successfully been applied for efficiently capturing detailed public opinion data (Ahnert et al., 2025; Cerina & Duch, 2023) – an application that could be transferred to open-ended responses. Other popular semi-automated classification approaches for open-ended responses, such as support vector machines or random forests (e.g., Haensch et al., 2022; Landesvatter, 2024), are less adaptable across different languages and often require substantial expertise, pre-processing, and training data coded by humans (Landesvatter, 2024). Since LLMs could potentially eliminate the need for these time- and expertise-intensive requirements, it is possible that they are an efficient alternative for classifying open-ended responses in survey research. While researchers have begun to explore this application of LLMs (Landesvatter, 2024; Mellon et al., 2024; Rytting et al., 2023) and were largely successful, most of these studies have focused on English-language responses, responses relating to non-complex topics, or on single LLMs and prompting strategies. It is thus unclear to what extent existing findings generalize to other LLMs, prompting strategies, languages, and more complex topics. Furthermore, research has raised concerns about the reproducibility of LLM-generated output due to their non-deterministic design (Barrie et al., 2024), an issue that can extend to the coding of open-ended responses when it comes to the reliability of the coding, for example when new survey data is available. Overall, the exact conditions of the applicability of LLMs for coding open-ended survey data and the quality of these classifications, also compared to more established methods, have yet to be understood.

In this project, we are the first to investigate to what extent different LLMs can be used to code non-English (German) open-ended responses on survey motivation given a predefined set of categories. We examine performance and reliability, and the dependency of these indicators on two factors – model selection and prompting approach. Specifically, we ask:

**RQ1:** Are there differences between LLMs regarding the performance and reliability of the coding?

**RQ2:** Are there differences between prompting approaches regarding the performance and reliability of LLM-based coding?

  **RQ2a:** Does providing detailed descriptions of categories improve the performance and reliability of the coding?

  **RQ2b:** To what extent does few-shot prompting impact the performance and reliability of the coding compared to zero-shot prompting?

  **RQ2c:** Does fine-tuning an LLM on a subset of pre-coded response data improve the performance and reliability of the coding?

To do so, we contrast proprietary and open-source LLMs – GPT-4o, Llama 3.2, and Mistral NeMo, which are the most capable multilingual models of their respective families to date. We compare their category assignments when using zero-shot prompting (i.e., not providing examples) with and without category descriptions and few-shot prompting (i.e., providing exemplary classifications), and fine-tuning (i.e., further training of the LLM), and evaluate them against the codings of human experts. We also discuss the LLMs' performance in contrast to other classification methods reported in previous studies. By comparing the use of different LLMs and prompting approaches for classifying open-ended survey responses in German, our study uniquely contributes to the growing body of research about the conditions under which LLMs can be efficiently, accurately, and reliably leveraged in survey research and about the impact of LLM use on data quality.

2. Background

There are three main types of approaches to coding open-ended survey responses: traditional human coding, supervised machine learning methods, and the still-emerging use of LLMs, each with distinct strengths and challenges. In this section, we review these methods, highlighting the potential of LLMs that yet needs to be explored.

In manual coding, human coders assign responses to predefined categories. While considered mostly accurate, this approach is time-intensive and costly, especially for large survey datasets or such with multiple open-ended questions (Landesvatter, 2024; Haensch et al., 2022). Costs are compounded when wanting to increase validity and reliability by having responses classified by several coders. These factors contribute to the sparseness of open-ended questions in survey instruments, despite such items allowing for deeper, authentic insights into how individuals think and act (Haensch et al., 2022).

Supervised methods attempt to address this resource-intensiveness by combining manual coding of a training dataset with machine learning algorithms, such as support vector machines (SVMs; Joachims, 2001) or gradient boosting (Schonlau & Couper, 2016). Applications to political (Grimmer & Stewart, 2013) and economic texts (Gentzkow et al., 2019) as well as other survey responses (Haensch et al. 2022; Schierholz & Schonlau, 2021) demonstrated their utility. But while these sophisticated approaches can somewhat reduce costs and time, they still require a substantial amount of human-coded data and expertise and computational resources for model training in order to achieve satisfactory results, making them inefficient. They also struggle with short open-ended survey responses, which often lack sufficient context. In addition, they are usually only trained for one specific language and topic, making them not easily transferable across studies and less feasible for multilingual studies.

Transformer-based models, such as BERT, are able to capture nuanced relationships in text due to their ability to generate contextual embeddings. This offers improved classification performance for open-ended survey questions (Meidinger & Aßenmacher, 2021; Gweon & Schonlau, 2024). For example, Schonlau et al. (2023) demonstrated BERT's effectiveness for coding German-language survey questions, such as the GLES "most important problem" question. However, applying BERT to survey data poses similar challenges as supervised methods, as open-ended responses are often too short to utilize the models' full potential, and fine-tuning them to the specific types of (con)text requires expertise and computational resources (e.g., Schonlau et al., 2023). While BERT is an *analytical* language model designed primarily for specific tasks like classification or entity recognition at the sentence or document level, modern-day *generative* large language models such as GPT-4 are designed to perform a broader range of generative and context-adaptive language processing tasks, including handling complex dialogs, summarization, and multilingual text generation. Such general-purpose LLMs thus show potential to address limitations of earlier approaches when applied to open-ended survey responses, like handling short responses when given only general information on their context, not necessarily requiring pre-coded data for training or fine-tuning, and being flexibly usable across languages. In addition, since off-the-shelf LLMs do not require large programming expertise, are relatively cost-effective, and can follow natural language instructions, they are more accessible to a broader group of survey researchers than other semi-automated methods. LLMs have brought promising advancements to labeling other types of social science text data, such as social media data and political texts, with studies finding that LLMs were at least on par or even outperformed supervised methods (Ahnert et al., 2025; Ornstein et al., 2024; Törnberg, 2024), making them applicable for substantive downstream analyses, like predicting public opinion (Cerina & Duch, 2023; Ahnert et al., 2025; Heseltine & Clemm von Hohenberg, 2024). Research specifically evaluating the applicability of LLMs for coding open-ended survey responses, however, continues to be scarce. In addition, LLMs' rapid evolution requires constant reevaluation of their precision and domain-specific applicability (Pangakis et al., 2023).

Rytting et al. (2023) tasked GPT-3 to code 7500 English open-ended responses on keyword descriptions of U.S. partisans into binary and ternary categories. The LLM-based coding matched the (poor) performance of human crowdworkers and experts in terms of inter-coder agreement. It also came close to the performance of a supervised approach while needing substantially fewer labelled examples. Mellon et al. (2024) come to similar conclusions when testing a larger and more recent variety of open- and closed-source LLMs for coding several thousand open-ended responses to the "most important issue" question in the British Election Study into 50 categories. Benchmarked against a trained human coder, LLMs' accuracy of classifications varied between and within model families. Compared to a range of supervised approaches, the general-purpose LLMs performed much better, with BERT-based methods still outperforming SVMs.

Using LLMs for coding open-ended survey responses thus appears like a promising method for survey researchers. However, these studies represent a best-case scenario of relatively easy tasks, as they cover English-language data about standard societal and political issues that are likely much-discussed in LLM training data and do not require much expertise for coding. Research on logical reasoning tasks suggests that LLMs tend to struggle with tasks that are comparably complex, but less commonly appearing in their training and alignment processes (McCoy et al., 2023). In addition, there is ample evidence that LLMs are biased against non-English language contexts in a variety of other tasks (e.g., Durmus et al., 2024; Johnson et al., 2022; Li et al., 2024; Wang et al., 2024). For example, Törnberg (2024) found that GPT-4 can be used for labeling non-English social media data, but Heseltine and Clemm von Hohenberg (2024) observed decreased speed and accuracy compared to English-language texts. Once again, these studies examined comparatively simple tasks, namely binary labeling of sentiment and political affiliation.

Beyond these limitations, there is competing evidence regarding specific LLM performance and prompting strategies: It is unclear whether all (families of) LLMs are equally suited for classifying open-ended responses. For example, most studies on using LLMs for coding social science text data investigated models of the GPT family, but came to conflicting conclusions regarding different model versions (e.g., Bosley et al., 2023 vs. Rytting et al., 2023 for GPT-3; Ornstein et al., 2024 vs. Heseltine and Clemm von

Hohenberg, 2024 and Törnberg, 2024, for GPT-4; Mellon et al., 2024 vs. Ahnert et al., 2025, for Llama). Considering proprietary vs. open-source model families, Mellon et al. (2024) found that the closed-source Claude models matched human coding best, followed by GPT-4, whereas Llama and PaLM performed much worse, and some other open-source LLM families were unable to complete the task at all.

Furthermore, existing research uses competing prompt designs. Some studies suggest zero-shot prompting (i.e., not providing examples for the labeling task, only the possible labels) is sufficient for labeling other types of short social science text data (Cerina & Duch, 2023), even in non-English languages (Törnberg, 2024). In contrast, studies applying LLMs specifically to open-ended survey responses used few-shot prompting (Mellon et al., 2024; Rytting et al., 2023). In this approach, the authors included the coding scheme along with three examples in the prompt, sometimes supplemented by detailed category descriptions. Halterman and Keith (2024) found that including more detailed definitions of the categories and positive as well as negative examples had a positive impact on labeling quality. However, Mellon et al. (2024) report that providing a full coding guide appeared to "distract" the LLMs. Finally, Mellon et al. (2024) suggest that fine-tuning, i.e., re-training LLMs on pre-labeled survey responses would likely further improve results. Ahnert et al. (2025) successfully used fine-tuning, albeit not for open-ended survey data.

Given this scarce and competing evidence, it remains unclear whether and which existing findings about the applicability of LLMs for coding open-ended survey responses generalize. In this study, we seek to close this gap by testing different LLMs and prompting strategies for multi-class, single-label classification of a more specific topic in German open-ended survey data.

## 3. Data and Methods

*Open-ended survey data & coding scheme*

In order to test the applicability of LLMs for coding German-language open-ended survey responses, we use data from a German probability-based mixed-mode panel, the GESIS Panel.pop Population Sample (Bosnjak et al., 2018, GESIS, 2024). Randomly sampled from municipal population registers, the panel includes over 5,000 respondents and covers the population of German-speaking permanent residents of Germany aged 18+. Participants are invited to the 20-minute survey waves bimonthly, receiving a prepaid incentive of five euros with every invitation. For the years 2014 to 2020, the survey includes an annual, non-mandatory open-ended question on survey motivation. There, the panelists are asked to give their most, second most, and third most important reason for participating in the panel on three separate lines (see Figure A1 in Appendix II for question wording). This questionnaire design leads to unidimensional answers usually containing only one category, making the item very favorable for coding (Haensch et al., 2022). Thus, while the response format should present an easy test case for LLMs, the specificity and complexity of the topic in terms of categorical dimensions, as well as the German language, present a harder task. The dataset contains a total of approximately 25,000 responses to the question on survey motivation across survey waves. For our study, we rely on a random sample of 20 percent of that data (5,072 responses) coded independently by two survey researchers (Cohen's kappa = 0.91, with remaining disagreements resolved by a more senior expert) based on a coding scheme for survey motivation adapted to the GESIS Panel.pop by the survey researchers (see Haensch et al., 2022 for details). The human codes are not necessarily required for employing LLMs (see below for a discussion of prompting approaches), but serve as a ground truth to compare the LLM-based classifications to. Indeed, when not fine-tuning an LLM, using it would require only a fraction of the human-coded examples necessary for training traditional supervised methods – for example, Haensch et al. (2022) used 5000 human-coded responses to train an SVM.

For the LLM-based classifications, we use the same coding scheme as was used by the human coders. It spans 22 categories, featuring both intrinsic, extrinsic, and survey-related reasons for motivation (Porst & von Briel, 1995; Haensch et al., 2022). It also includes catch-all categories: *No reason* captures explicit statements of not having a reason for participation, "don't know"s, as well as non-meaningful fillers such as "???". In contrast, *Other* contains meaningful statements that cannot be assigned to any other category. For English translations of the categories, see Figure 1. A more detailed coding scheme with definitions and examples for all categories and their groups can be found in Appendix I.

*LLM selection & configuration*

We test and compare powerful and popular LLMs of three different model families that are state-of-the-art at the time of writing. Models of one of the industry leaders, OpenAI, are popularly used by the public and researchers without large computational expertise due to their user-friendly accessibility. Despite OpenAI's lack of transparency and reproducibility as a proprietary provider (Palmer et al., 2023), it thus is reasonable to include one of their models in our research as a realistic use case. GPT-4o (GPT henceforth) is OpenAI's flagship model at the time of writing, which, according to the developers, features considerable improvements in non-English languages over earlier versions, while being more time- and cost-efficient (OpenAI, 2024a, 2024b). It is also supposed to be more capable of domain-specific or complex tasks and detailed labeling.

In line with calls for accessible and reproducible AI research (e.g., Spirling, 2023; Weber & Reichardt, 2024), we also test two open-source LLMs. These are downloaded and run locally, ensuring sensitive data remains private and is not shared with third parties. This is crucial as open-ended responses may inadvertently contain personal information, such as addresses, risking re-identification.[1] Running LLMs locally also ensures reproducibility by using stable model versions, unaffected by updates to cloud-based APIs (Spirling, 2023). Llama-3.2-3B-Instruct (Llama henceforth) is the more capable of the two multilingual LLMs of Meta's Llama 3.2 suite, the most recent and powerful one at the time of writing (Meta, 2024a, 2024b). While the open-source suite also features larger models (11B and 90B), those are not optimized for multilingual dialog and not available in Europe, making them infeasible for the project at hand and international survey research more broadly. Mistral-NeMo-Instruct-2407 (Mistral henceforth) is the most recent multilingual model by the European open-source developers Mistral. It is specifically designed for global, multilingual applications (MistralAI, 2024a) and supposed to be particularly strong in, among other languages, German. We access these models via the Huggingface platform (Meta, 2024b, MistralAI, 2024b).

---

[1] To ensure a similar level of privacy for the proprietary GPT model, we (fine-tune and) run it on AzureOpenAI, which provides private instances of GPT models on European servers.

To investigate the exact conditions under which LLMs can be used to code German open-ended survey responses, we employ different approaches.

**Zero-shot prompting:** In the least supervised approach, we simply ask the LLMs to classify the open-ended responses without any additional information apart from the coding scheme (i.e., no examples or definitions of responses belonging to the specific categories).

**Zero-shot prompting with category descriptions:** Along with the coding scheme, we provide the LLMs with definitions for each category.

**Few-shot prompting:** In few-shot prompting, an LLM is given a few examples to guide its output along with the coding scheme, providing an efficient alternative to training the LLM with task-specific data. To test how few-shot prompting impacts the performance of LLMs for open-ended response classification, we provide the LLMs with one example response per category (so 22 examples in total) in the prompt. The examples are randomly selected from the examples featured in the coding scheme, containing actual answers featured in the dataset of responses to be classified. They are presented in random order in the prompt. The examples are not removed from the classification dataset.

**Fine-tuning:** Fine-tuning involves further training the model on a smaller, domain-specific dataset to improve its performance on particular tasks. While less efficient because of the need for more human-coded training examples, fine-tuned LLMs might yield more accurate results than using LLMs out-of-the-box. Exploring whether fine-tuning a model on humanly pre-coded response data thus helps understand LLMs' potential in classifying open-ended responses.

However, depending on the LLM, fine-tuning requires even more extensive computing resources. This is not only a limitation for practitioners, but also for our test case. We therefore select only GPT-4o for fine-tuning, due to its straightforward and easily available fine-tuning services, making it a likely choice for researchers wishing to employ this approach. We fine-tune the LLM by splitting the dataset into a training and a test subset. As is common for fine-tuning tasks, we randomly select 80 percent of responses of each category based on the human classification (4048 in total)[2] for training the LLM before asking it to classify the remaining 1024 responses using the zero-shot prompt. Results for the fine-tuned approach thus reflect the LLMs' performance on the test set alone. We specify four epochs[3] for fine-tuning, i.e., four iterations through the training data, and use default values for batch size (the number of examples used in a single training pass; around 0.2% of the training dataset, ten in our case) and learning rate (rate at which the LLM updates its weights (i.e., internal settings) based on the new data, balancing between learning too slowly, risking inefficiency, and too quickly, risking instability). Appendix II.6 reports the loss and token accuracy curves of the fine-tuning process.

Since we want to maximize reliability and the task of coding responses according to a set of predefined categories does not require creativity but consistency, we set the LLM temperature to 0, thereby flattening the LLM's underlying probability function to produce more deterministic outputs. For best comparability, we use the same temperature for all models, leaving all other parameters at model default.

---

[2] We train the LLM with the zero-shot prompt including the responses in their raw form as input, not correcting any spelling mistakes or similar. As output, we use the desired completion format (see prompt design).

[3] Using four epochs for fine-tuning is a deliberate choice, balancing generalization and task-specific adaptation. While this is not the default, it represents a compromise between the lower range typically sufficient when using validation sets (i.e., 1–2 epochs) and OpenAI's recommendation to increase the number of epochs for tasks with a small set of ideal outputs, such as classification.

```
You are a survey expert classifying open-ended responses to the question why individuals
participate in a survey. Assign these reasons for participating to exactly one of the following
categories.

The categories are:
INTEREST: [Description]
CURIOSITY: [Description]
LEARNING: [Description]
TELL OPINION: [Description]
INFLUENCE: [Description]
INCENTIVE: [Description]
FUN: [Description]
ROUTINE: [Description]
DUTIFULNESS: [Description]
HELP SCIENCE: [Description]
HELP POLITICIANS: [Description]
HELP SOCIETY: [Description]
HELP, NOT FURTHER SPECIFIED: [Description]
BREVITY: [Description]
ANONYMITY: [Description]
PROFESSIONALISM: [Description]
RECRUITMENT: [Description]
RECRUITER: [Description]
OTHER SURVEY CHARACTERISTICS: [Description]
IMPORTANCE IN GENERAL: [Description]
OTHER: [Description]
NO REASON: [Description]

Make your best guess, even if it is hard.
Respond in the following format: Reason for participating | CATEGORY.
Do not give an explanation for your classification, but return only the reason for participating and
your classification.

Examples:
[Example reason | CATEGORY 1]
[Example reason | CATEGORY 2]
[...]
[Example reason | CATEGORY 22]

Classify the following reason for participating:

[open-ended response]
```

*Figure 1: English translation of prompt used for LLM-based classifications of the open-ended survey question. Categories and, in the detailed approach, descriptions (green font) were randomized across individual queries. In the few-shot approach, examples (blue font) were randomly selected, the selection being held constant, but presented in random order across queries. For details of descriptions and examples used, see Appendix I.*

*Prompt design*

We tell LLMs to impersonate a survey expert classifying open-ended responses and instruct them to assign each response to exactly one category. The order of categories (and their descriptions in the detailed approach, and examples in the few-shot approach, respectively) is randomized in each prompt to avoid any biases due to order effects (Brand et al., 2023; Pezeshkpour & Hruschka, 2024). To minimize missing values, we ask the LLMs to make a best guess in difficult cases. We instruct the LLMs to report the response along with its classification. Finally, to avoid unnecessarily long answers, we ask the LLMs not to justify their response (as especially Mistral has been found to do previously, see e.g., von der Heyde et al., 2025), but do not specify a maximum output length. Figure 1 shows an English translation of the prompt. In line with the language of the responses they are being asked to classify, we prompt the LLMs in German, including the instructions and coding scheme. The original German version of the prompt, as used in the study, can be found in Appendix II (Figure A2).

We prompt each survey response separately and with refreshed LLM memory, to ensure that responses are classified independently of one another. We therefore specify the task directly in the main prompt (not the system prompt), thereby repeating the task for every open-ended response to be classified. Before we feed the full dataset to the LLMs, we test each LLM with only 15 responses to determine its general capacity to fulfill the task. We run each query twice per LLM to be able to evaluate its reliability. All data is generated in November 2024, except the classifications obtained from the fine-tuned version of GPT, which is generated in January 2025.

*Analysis*

We extract each LLMs' classifications of the open-ended responses and analyze their performance and the resulting descriptive distributions. Benchmarking against the human-generated classifications, we analyze the LLMs' classification performance overall and per category. Because our case is one of multiclass-classification and the benchmark categories are unevenly distributed (see Figure 4), we use macro F1 scores[4] as our primary overall performance metric (Hand et al., 2024). In imbalanced datasets, regular F1 scores can be misleading if an LLM tends to assign the modal category. Macro F1 addresses this by averaging across the per-category F1 scores, giving equal weight to minority categories.

If an LLM failed to classify a response to exactly one category (i.e., it did not assign a category or assigned more than one category), the output is recorded as missing (i.e., an explicit category called "NA") but retained for the analysis. This approach avoids artificially inflating the F1 scores for categories where most responses were not classified, but the remainder classified correctly,[5] and allows us to investigate the reliability of missing classifications. To facilitate comparison to other studies and classification methods, we report additional metrics (weighted F1, accuracy, intraclass correlation coefficients, Cohen's kappa) in Appendix II (Table A1). Since Haensch et al. (2022) previously tested an SVM on the same data, we are able to compare LLM performance to that of a supervised approach without explicitly having to employ that approach ourselves (see the discussion). To do so, we calculate the median F1 score as the unweighted median across categories. We then compare the distribution of coding scheme categories across LLMs and prompting approaches and to the distribution of the human-coded benchmark data. We also report the frequency and categorical distribution of the responses each LLM fails to classify as well as the reason for failure (see Appendix II.3). For all analyses, we rely on the first iteration of classifications per LLM and prompting approach, independent of whether this iteration exhibited better or worse performance than the second one, in order not to bias our results by selecting on performance.

---

[4] Generally, F1 scores range from 0 to 1, with higher values indicating better predictive performance. For a more detailed description, see Appendix II.4.

[5] The overall macro F1 scores exclude "missing" as an assigned category, as it is not meaningful or valid.

To assess the LLMs' reliability, we calculate the ICC for two-way agreement between the two iterations of classifications per LLM and prompting approach.

Data (pre-)processing, classification (for GPT), and analyses are conducted in R (version 4.3.2, R Core Team, 2024), especially using the packages *AzureAuth* (Ooi et al., 2019), *caret* (Kuhn, 2008), *irr* (Gamer et al., 2019), and *tidyverse* (Wickham et al., 2019). Classifications from Llama and Mistral are obtained using Python, especially using the packages *accelerate (Accelerate, n.d.), huggingface_hub (Hub Client Library, n.d.)*, *pandas* (McKinney 2010), *PyTorch* (Paszke et al., 2019), *tqdm* (Casper da Costa-Luis et al., 2024), and *transformers* (Wolf et al., 2020).

## 4. Results

### 4.1. RQ1: Differences between LLMs

*Performance*

We first compare differences between LLMs in classification performance overall (macro F1) and per category (F1). Across prompting approaches, classification performance is much better when using GPT than when using Mistral, which still has a slight edge over Llama (see Figure 2). GPT performance also fluctuates much less between prompting approaches (macro F1 around 0.7 for the three approaches that were examined for all three LLMs). Nevertheless, even using the best-performing prompting approach for an open-source LLM does not come near the GPT performance. Similar patterns emerge when considering other performance metrics (see Table A1).

All LLMs examined exhibit approximately the same performance patterns across categories (see Figure 3). They perform exceptionally well on the categories *incentive, interest,* and *fun* (macro F1 around 0.9), as well as on *anonymity, routine,* and *tell opinion*, and exceptionally poor (macro F1 between 0.02 and 0.3) on *no reason, non-identifiable/other,* and *other survey characteristics*. The LLMs thus perform very well on the three categories most commonly defined by the human coders, but not on the next two most common categories, which are non-substantive catch-all categories. For the remaining categories, performance tends to decrease along with frequency of occurrence. The overall pattern is mirrored across types of reasons (extrinsic, intrinsic, survey-related): GPT's performance tends to be better than that of Llama and Mistral, which improves with few-shot prompting. There are some cases that stand out, which help explain the overall performative edge GPT has over the open-source models. GPT outperforms Llama and Mistral especially in *tell opinion, routine, importance in general, influence, dutifulness, curiosity,* and *professionalism,* and to a lesser extent also in the *help* categories, although Llama and especially Mistral improve under few-shot prompting.[6]

---

[6] For "learning" under zero-shot prompting (with and without definitions) and for "recruiter" when prompting with descriptions, no macro F1 scores can be calculated for Llama, indicating that there were no true positives, no false positives (i.e., the LLM did not assign any of the responses to that

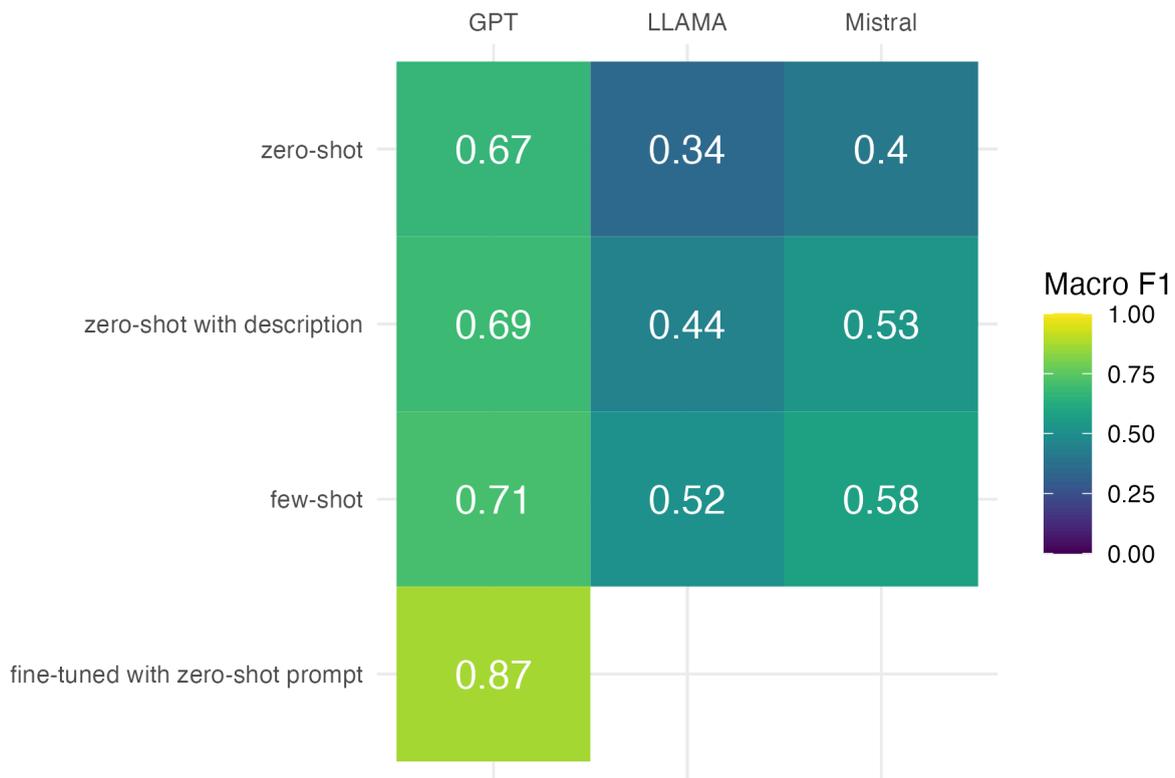

*Figure 2: Macro F1 scores by LLM and prompting approach.*

*Distributions*

The differences in LLMs' classification performance across categories result in different frequencies of categories (see Figure 4), although the overall shape of the distribution is similar to the human-coded benchmark. While the LLMs' good performance on classifying *incentive, interest,* and *fun* leads to the proportion of responses in these categories being close to the human benchmark, their poor performance on other categories manifests in substantially lower proportions than the human data would suggest. This includes *non-identifiable/other,* which is among the five most frequently identified categories according to the human coders. Llama and Mistral additionally assign too few cases to *no reason*, but code more responses as *curiosity* than both humans and GPT, where they also perform worse in terms of F1 scores. Conversely, the proportion of responses assigned to *tell opinion* tends to be lower when using Llama. In contrast, Mistral assigns disproportionately many responses to *tell opinion*, and, to a smaller degree, to *help society, help science,* and *recruitment* – the categories where GPT tends to outperform.

The proportion of missing (including ambiguous) assignments is (initially) higher for the open-source models than for GPT. Just as with performance and overall distribution, GPT is also less sensitive to prompting approaches than other LLMs when it comes to missing classifications, whereas the performance of Llama and Mistral depends on the prompting approach. Both open-source models eventually return better results than GPT when considering the amount of missing classifications. When using GPT, missing category) for these categories. The same is true for Mistral for the category "no reason" under zero-shot prompting.

classifications occur almost exclusively for responses labeled as *no reason* by human coders (Figure A3), with over 60 percent of responses lacking a classification. In contrast, missing classifications are more evenly distributed across all categories when using Llama (which also misses assignments for close to 60 percent of *no reason* responses) or Mistral. This partly helps explain the poor classification performance for the *no reason* category; however, missing values cannot account for the poor performance on other categories (see Appendix II.5 for full confusion matrices; and Appendix II.3 for F1 scores when omitting missing values).

*Reliability*

Turning to reliability of the classifications, Mistral's output is identical across the two iterations, proving to be the only LLM tested where setting the temperature to zero and setting a seed actually results in the desired behavior – returning identical and therefore reliable output.[7] Yet, the other two LLMs also exhibit high reliability (ICC > 0.93, see Table 1). There are only minimal differences, with GPT being slightly more reliable than Llama.

| Approach | GPT-4o | Llama 3.2 | Mistral NeMo |
|---:|:---:|:---:|:---:|
| **zero-shot** | 0.99 | 0.95 | 1.00 |
| **with descriptions** | 0.99 | 0.94 | 1.00 |
| **few-shot** | 0.99 | 0.95 | 1.00 |
| **fine-tuned** | 0.99 | | |

*Table 1: ICC (two-way agreement) between two rounds of coding per LLM and prompting approach.*

In sum, there are differences between LLMs in terms of performance and, to a lesser extent, reliability when coding German open-ended survey responses. Disregarding prompting approaches, using GPT results in higher classification performance than using Llama or Mistral, but performance under GPT is still subpar, both in absolute terms and relative to other methods (e.g., Haensch et al., 2022) when not using fine-tuning (see below). While all LLMs exhibit high reliability across iterations, Mistral has a slight edge, reproducing the exact same classifications.

---

[7] It is possible that the discrepancies in the other two LLMs are caused by the temperature not being implemented as zero by the LLMs (despite setting it as such), but a very small number, for mathematical reasons. Temperature is a normalization parameter for the LLM's underlying softmax function; setting it to zero would result in division by zero. We briefly discuss the implications of this in the next section.

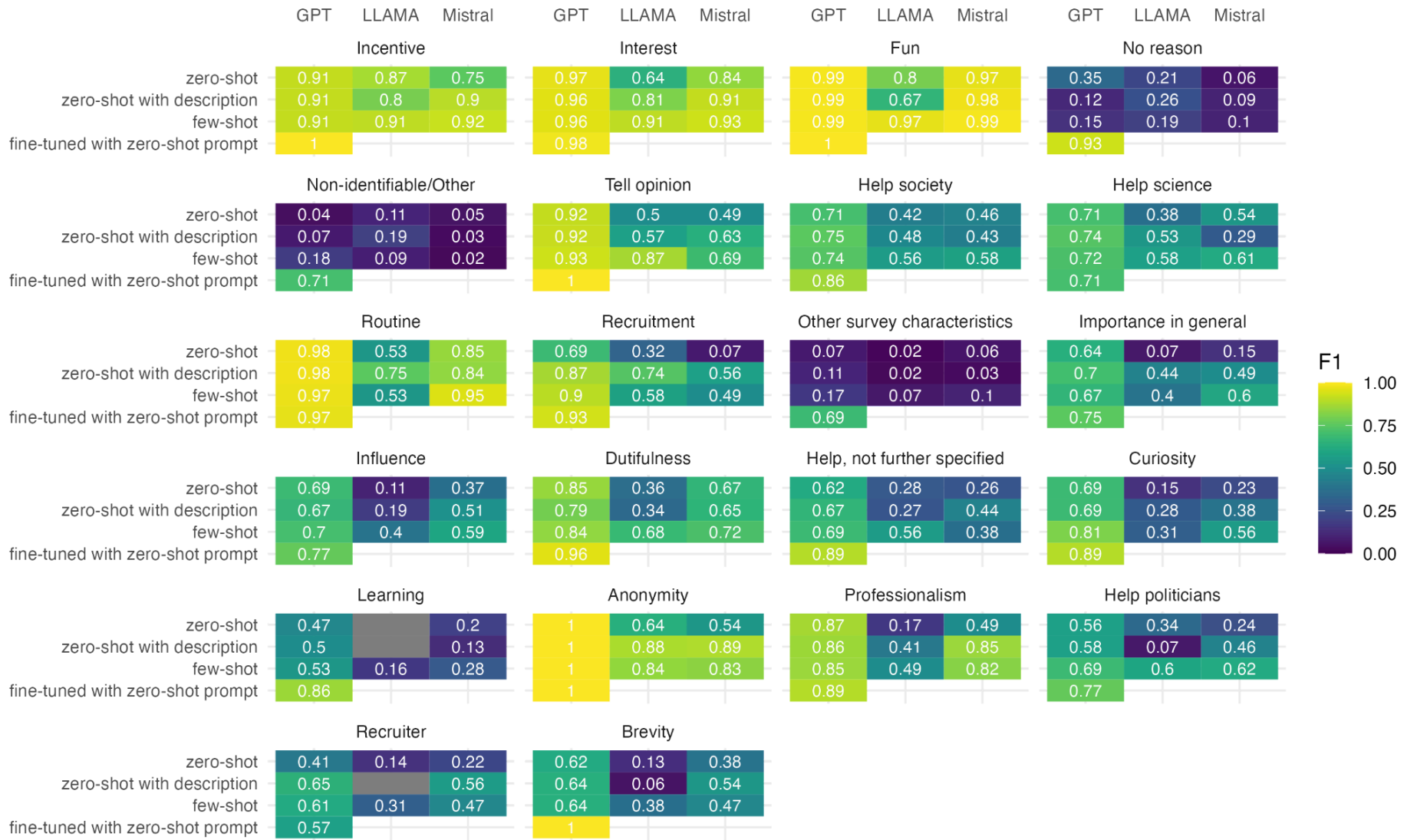

Figure 3: Per-category F1 scores by LLM and prompting approach.

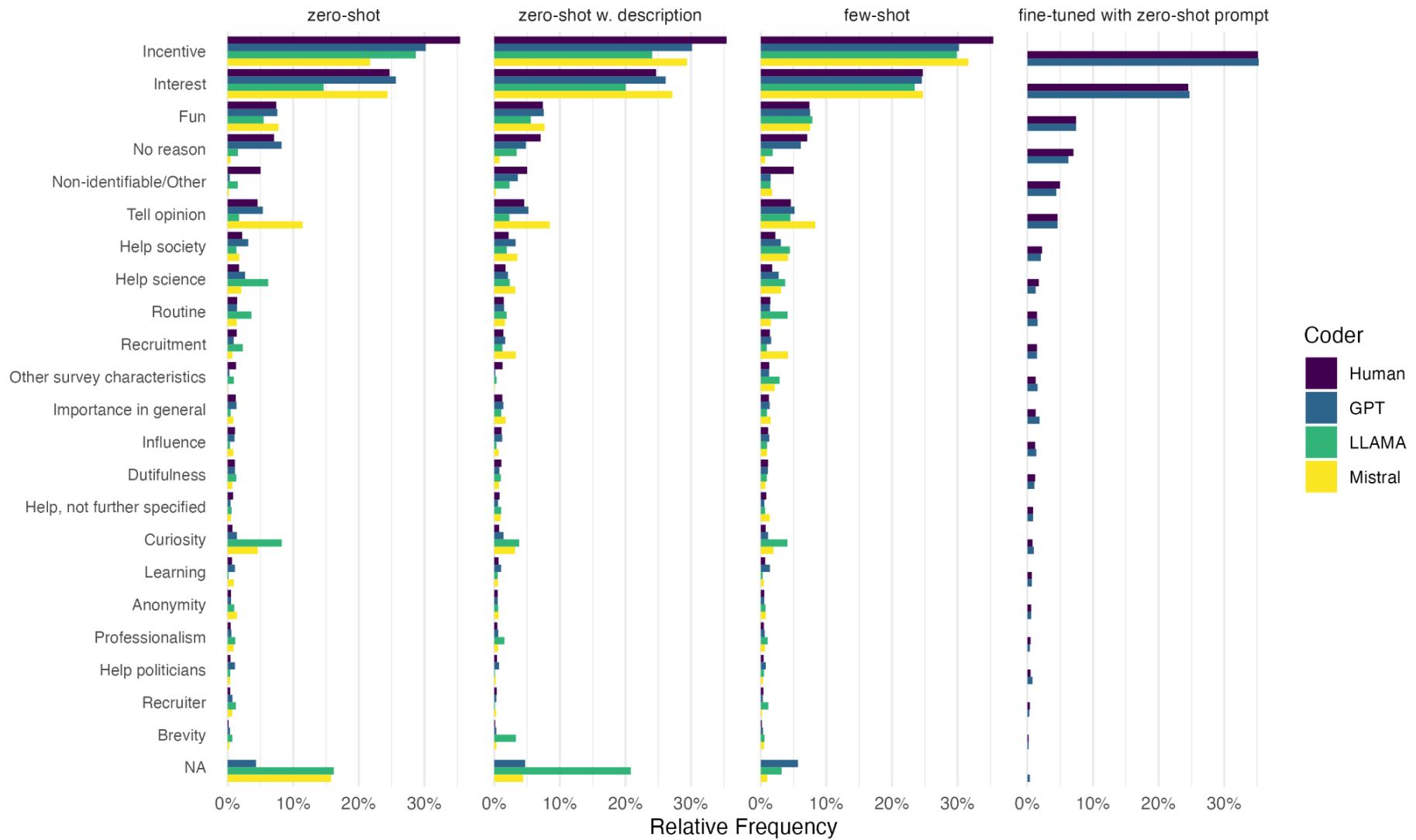

Figure 4: Distribution of coding categories by LLM and prompting approach.
n=5072 for zero-shot (with and without description) and few-shot prompting, n=1024 for fine-tuned prompting.

## 4.2. RQ2: Differences between prompting approaches

*Performance*

When comparing differences in classification performance between prompting approaches across LLMs, performance is best for few-shot prompting and worst for zero-shot prompting in terms of macro F1. However, the size of the difference depends on the LLM used. There is a strong improvement in performance from zero-shot prompting to few-shot prompting when using the open-source models – for both models, there is a 0.18 difference in F1 macro scores, see Figure 2. The same pattern emerges when considering other performance metrics (Table A1), and when investigating classification performance per category (Figure 3). However, for singular combinations of LLM used and category classified, performance is worse when providing the LLM with descriptions than when using simple zero-shot prompting (e.g., *fun, no reason, help science*), or when providing examples relative to providing descriptions (e.g., *recruitment, anonymity, non-identifiable/other*). This is more often so for the open-source LLMs than for GPT.

Most notably, GPT's performance drastically improves when employing fine-tuning, achieving a macro F1 of 0.87 – a 16 point difference over few-shot prompting and a satisfactory level in general. This jump can largely be attributed to much improved classification in the non-substantive categories. For other categories, a mixed picture emerges, with large improvements for six categories, but minor improvements for the remainder – in part because few-shot prompting already led to high levels of performance.

*Distributions*

Although all prompting approaches examined approximately result in very similar distributions of categories, few-shot prompting tends to approximate the distribution of the human-coded data best (Figure 4). This is especially the case for *interest* and *tell opinion*. Large differences remain especially for *non-identifiable/other, help science,* and *help society*. Few-shot prompting also results in substantially fewer responses that were not coded successfully, with a reduction of almost four fifths for Mistral. As a consequence, there are almost no missing classifications under few-shot prompting, except for *no reason* (Figure A3). Fine-tuning results in a distribution that perfectly matches the human classifications, with only four classifications missing in total (all belonging to the *no reason* category).

*Reliability*

All LLMs exhibit high reliability (>0.93) regardless of approach when considering ICCs (see Table 1). Mistral is completely deterministic in all approaches, GPT is consistently very reliable across approaches, including fine-tuning, and Llama is slightly less reliable when provided with descriptions.

To summarize, the prompting approach used does make a difference in terms of performance, but not so much in terms of reliability of coding German open-ended survey responses. Providing detailed descriptions of categories tends to improve classification performance over zero-shot prompting, and few-shot prompting further improves it, especially for the open-source LLMs. Fine-tuning leads to the best overall performance and the largest improvement compared to other prompting approaches when using GPT. Reliability is high regardless of the prompting approach used.

## 5. Summary and Discussion

In our study, we assessed the performance and reliability of three powerful, multilingual LLMs (GPT-4o, Llama 3.2, and Mistral NeMo), when classifying German open-ended survey responses on a specific and complex topic given a pre-defined coding scheme. We also investigated differences depending on the prompting approach used. Overall, performance differed greatly between LLMs, and only a fine-tuned LLM achieved satisfactory levels of predictive performance (macro F1 of 0.87). In general, GPT performed best, and, disregarding fine-tuning, few-shot prompting led to the second-best performance (macro F1 of 0.71 for GPT), echoing the findings of previous studies on English data on less specific topics (Haltermann & Keith, 2024; Mellon et al., 2024). Performance differences between prompting approaches were conditional on the LLM used – the prompting approach was not as important when using GPT, but made a big difference for other LLMs, especially Mistral. While the LLMs correctly identified most of the responses belonging to the most frequently occuring (and most easily identifiable) reasons, they struggled with non-substantive catch-all categories. Limitations in performance in these categories may arise because human coders classified responses such as "don't know", "xxx", and blank responses as *no reason*. The LLMs often failed to categorize such data, instead treating it as if it contained no response. This is problematic for open-ended response classification more broadly. Responses belonging to such categories are quite common regardless of question topic, as many survey respondents lack the time or motivation to respond to open-ended questions, either giving non-substantive or nonsensical responses that practically correspond to item-nonresponse (Krosnick and Presser, 2010). In our case, LLMs' unequal classification performance across different categories of reasons for survey participation results in different categorical distributions when not using fine-tuning. Such discrepancies could also have consequences for further inferential analyses of the coded data. Thus, LLM-coded open-ended responses could paint a very different picture of the concept being measured by a survey item than human coding would.

  Our study shows that using off-the-shelf (i.e., non-fine-tuned) LLMs is not necessarily superior to other computational methods for coding open-ended responses. Comparing our results to those of Haensch et al. (2022), who used an SVM on the same data, even few-shot performance proved to be below expectations when going beyond the most obvious and common categories (median F1 0.83 vs. 0.72 at best). This is at odds with Mellon et al.'s (2024) findings regarding English-language survey responses on a more common topic: Although that study also reported that GPT models were superior to Llama models, it also found that the LLMs, when provided with the full coding scheme including descriptions *and* examples for over 50 categories, were much better at classifying British responses to the commonly discussed "most important problem" question than established supervised approaches, including BERT and SVMs. Rytting et al. (2023) came to similar conclusions even for the by now outdated GPT-3 under few-shot prompting, albeit for a task with only three categories. It thus appears that the applicability of LLMs for coding open-ended responses depends not just on the LLM and prompting approach used, but also on the topic (in terms of specificity and categorical complexity) and possibly language of the responses.

However, as our findings show, LLMs have the potential to match or even outperform other methods when fine-tuned. Using the zero-shot prompt on the fine-tuned GPT achieved a macro F1 of 0.87 (median F1 0.88), with dramatic improvements for non-substantive responses. This resulted in perfectly matched distributions between human and LLM-coded responses and virtually no missing classifications. Although this confirms speculations in terms of improved effectiveness over off-the-shelf usage (Mellon et al., 2024), it does not yet fulfill the hopes of being a resource-efficient alternative to established methods. This is because fine-tuning LLMs requires a sufficiently large set of human-coded benchmark data and more computational resources and expertise, similar to established methods, with which researchers are often more familiar. In addition, such established methods usually do not require payment, whereas proprietary LLMs (potentially requiring less programming expertise if providing user-friendly interfaces for fine-tuning) do. Additionally, this approach, as all others, relies on a pre-defined coding scheme, which may not readily exist for all open-ended questions practitioners might want to have classified.

While all three models we examined were very reliable in their classifications across two iterations, only Mistral showed the desired behavior of identical output when setting the model temperature to zero and setting a seed. The possibility that setting the temperature to the least probabilistic setting does not actually guarantee deterministic behavior can be unintuitive for survey researchers not familiar with LLMs in-depth, potentially risking a false sense of confidence. Yet, even the deviating LLMs in our study were more reliable than previous studies suggested (e.g., Heseltine & Clemm von Hohenberg, 2024), making resolvement by human coders (which, in the aforementioned study, did not exhibit higher agreement) obsolete. However, reproducibility over longer periods of time, e.g., for several survey waves featuring the same open-ended item, is not guaranteed when using non-local models, due to them being subject to change or deprecation. This highlights the need for regular validation with humans in the loop (see also Weber & Reichardt, 2024), even under high performance (which we only observed for the fine-tuned approach).

Our results also highlight the trade-offs between proprietary and open-source LLMs in terms of cost, privacy, reliability, and performance. Using open-source models such as Llama and Mistral, available on platforms such as Huggingface, are free to use and can be run locally, ensuring privacy and reproducibility by avoiding third-party servers and model updates. However, running them requires considerable computing resources and expertise, which not all researchers may have access to. In contrast, proprietary models like GPT, while user-friendly, incur costs per token (i.e., input and output length), which can be high for large datasets or complex instructions.[8] In our case, open-source LLMs underperformed compared to proprietary ones in coding open-ended responses, and fine-tuning a GPT model was the most successful approach. Finally, the speed of advancement of LLMs presents researchers with the challenge of working towards a moving target, where working with reliable and reproducible model versions may not present the state of the art.

---

[8] Per iteration through the dataset, we spent about EUR 10 for zero-shot prompting with GPT, EUR 20 for zero-shot prompting with descriptions, EUR 15 for few-shot prompting, and EUR 60 for fine-tuning and zero-shot prompting the fine-tuned model. Considering that in our case, inference took between 2 and 6 hours per iteration depending on the prompting approaches when self-hosting Llama 3 and Mistral on a A100 GPU, if renting such resources cost around EUR 2 per hour, this would result in an estimated cost of EUR 4-12 per iteration. Fine-tuning on the same dataset might require 4-6 hours, adding a one-time cost of about EUR 10-15. However, precise cost estimation is difficult due to variability in model size, hardware availability, batch optimization, and additional engineering overhead.

Our work gives rise to some further considerations and possible improvements. First, more experiments with different prompting strategies (Schulhoff et al., 2024) could be explored to see whether fine-tuned performance can be neared or made more cost-effective. For example, even more explicit instructions emphasizing the importance of always assigning a category and exactly one category might improve results especially for non-substantive responses. Researchers could also investigate whether breaking down the task into a two-step process would reduce its complexity by shortening the coding scheme information to be processed per prompt, and lead to more satisfactory results. In this prompt-chaining approach, the LLM could first be asked whether a specific category would be suitable for an answer. After having iterated across all possible categories in the coding scheme, the LLM could then be asked for the best-suited category from among the set of those it identified as suitable. Such an approach would allow for more examples per category in the first step without negatively impacting the LLM's context capacity (see, e.g., Mellon et al., 2024), thereby possibly improving performance. For fine-tuning, future research should focus on systematic experiments to identify the minimum amount of human-coded data needed for effective performance, balancing resource efficiency with accuracy. Additionally, LLMs' inner workings, including how they process different languages relative to one another, are somewhat opaque and not always consistent (see e.g., Zhang et al., 2023) – they might be better aligned to follow English instructions and coding schemes regardless of the language of the text to be classified. It is thus possible that LLMs perform better on non-English text classification when instructed in English, i.e., when only the survey response is in the native language. This would allow for simultaneous coding and translation of open-ended survey responses (Heseltine & Clemm von Hohenberg, 2024). Future research could investigate this by employing the English translation of our prompt.

Second, our study focused on the performance and reliability of LLM-coded open-ended survey responses, without investigating the impact of the method on the findings of substantive analyses. Replications of earlier substantive analyses that used more established classification methods with a fine-tuned LLM could complement our research. As part of such an analysis, taking into account uncertainty could shed light on whether distributional differences between LLM-based and human classifications are systematic. This could be done by analyzing the LLM's internal token probabilities (i.e., the probability with which the output is chosen), choosing the majority category after multiple iterations using an LLM's default temperature, or by directly asking the LLM for its certainty in a specific label (e.g., Tian et al., 2023). However, if human coders are inconsistent, models may be unfairly penalized, leading to deceptively low accuracy metrics. Even high inter-rater agreement (e.g., Cohen's kappa) can mask systematic errors made consistently by humans and mimicked by the model. In addition, human coders could, consciously or unconsciously, introduce biases based on their positionalities and stereotypes, which also may affect the coding and the evaluation metrics of the LLM.

Relatedly, LLMs might detect patterns or nuances humans do not, especially when not constrained by a fixed coding scheme. Using LLMs for unsupervised approaches, such as topic modeling (e.g., Ornstein et al., 2024), could address this concern while also making the ex-ante development of coding schemes for new survey items obsolete (Mellon et al., 2024), further increasing efficiency compared to supervised methods. However, results from unsupervised approaches are challenging to evaluate due to the absence of ground truth labels and because the interpretations of discovered patterns are often subjective (Pham et al. 2024). In addition, even if humans are subjective, the large discrepancy between human and LLM-based codes in our study suggests the latter are systematically mistaken (see

Fröhling et al., 2024, for a suggestion for diversifying LLM annotation). Depending on the complexity of the response data, it thus appears that off-the-shelf LLMs are not able to capture human reasoning as expressed in open-ended survey responses when not fine-tuned with human-coded benchmark data.

## 6. Conclusion

At a time when LLMs are revolutionizing survey research, there have been high hopes for their applicability to coding open-ended survey responses. Other studies have demonstrated singular LLMs' promising potential when tasked to code responses in comparatively easy contexts. However, we have shown that these findings do not necessarily generalize to other topically or linguistically more complex contexts: There is no one-size-fits-all kinds of open-ended response data regarding the LLM or prompting approach used. Even for the same data, using just any LLM for coding does not work equally well, nor does it work automatically without humans in the loop. Instead, it requires careful prompt engineering or, even better, fine-tuning with data pre-coded by humans. When coding German open-ended responses on a very specific topic with a complex classification scheme, LLM performance is generally low and differs greatly between LLMs. In addition, differences in prompting approaches are conditional on the LLM used. Comparing GPT, Llama and Mistral, using a fine-tuned version of GPT resulted in the highest classification performance. When not fine-tuning, however, classification quality is low compared to other, "easier" application contexts (English-language responses to more common survey items) and other classification methods (supervised machine learning models). LLMs may thus be an effective and possibly efficient alternative in such easier settings, provided a pre-defined coding scheme exists – but success is not guaranteed. Our results indicate that the specific LLM and prompting approach to be used for coding open-ended responses needs to be thoroughly validated before deployment. For more difficult (con)texts, fine-tuning on human-coded data increases the chances of success. Thus, as of now, humans still need to be in the loop for the coding and analysis of open-ended survey responses.

## References

Ahnert, G., Pellert, M., Garcia, D., & Strohmaier, M. (2025). Extracting Affect Aggregates from Longitudinal Social Media Data with Temporal Adapters for Large Language Models. *Proceedings of the International AAAI Conference on Web and Social Media*, *19*(1), 15–36. https://doi.org/10.1609/icwsm.v19i1.35801

Argyle, L. P., Busby, E. C., Fulda, N., Gubler, J. R., Rytting, C., & Wingate, D. (2023). Out of One, Many: Using Language Models to Simulate Human Samples. *Political Analysis*, 1–15. https://doi.org/10.1017/pan.2023.2

Barrie, C., Palmer, A., & Spirling, A. (2024). *Replication for Language Models*. https://drive.google.com/file/d/1wNDIkMZfAGoh4Oaojrgll9SPg3eT-YXz/view

Bosley, M., Jacobs-Harukawa, M., Licht, H., & Hoyle, A. (2023). *Do we still need BERT in the age of GPT? Comparing the benefits of domain-adaptation and in-context-learning* *approaches to using LLMs for Political Science Research*.

Bosnjak, M., Dannwolf, T., Enderle, T., Schauer, I., Struminskaya, B., Tanner, A., & Weyandt, K. W. (2018). Establishing an Open Probability-Based Mixed-Mode Panel of the General Population in Germany: The GESIS Panel. *Social Science Computer Review*, *36*(1), 103–115. https://doi.org/10.1177/0894439317697949

Brand, J., Israeli, A., & Ngwe, D. (2023). *Using GPT for Market Research* (SSRN Scholarly Paper No. 4395751). https://papers.ssrn.com/abstract=4395751

Casper da Costa-Luis, Stephen Karl Larroque, Kyle Altendorf, Hadrien Mary, richardsheridan, Mikhail Korobov, Noam Yorav-Raphael, Ivan Ivanov, Marcel Bargull, Nishant Rodrigues, Shawn, Mikhail Dektyarev, Michał Górny, mjstevens777, Matthew D. Pagel, Martin Zugnoni, JC, CrazyPython, Charles Newey, … Jack McCracken. (2024). *tqdm: A fast, Extensible Progress Bar for Python and CLI* (Version v4.67.1) [Computer software]. Zenodo. https://doi.org/10.5281/ZENODO.595120

Cerina, R., & Duch, R. (2023). *Artificially Intelligent Opinion Polling* (No. arXiv:2309.06029). arXiv. http://arxiv.org/abs/2309.06029

Cuevas, A., Brown, E. M., Scurrell, J. V., Entenmann, J., & Daepp, M. I. G. (2023). *Automated Interviewer or Augmented Survey? Collecting Social Data with Large Language Models* (No. arXiv:2309.10187). arXiv. http://arxiv.org/abs/2309.10187

Durmus, E., Nguyen, K., Liao, T., Schiefer, N., Askell, A., Bakhtin, A., Chen, C., Hatfield-Dodds, Z., Hernandez, D., Joseph, N., Lovitt, L., McCandlish, S., Sikder, O., Tamkin, A., Thamkul, J., Kaplan, J., Clark, J., & Ganguli, D. (2024). Towards Measuring the Representation of Subjective Global Opinions in Language Models. *First Conference on Language Modeling*. https://openreview.net/forum?id=zl16jLb91v

Fröhling, L., Demartini, G., & Assenmacher, D. (2024). *Personas with Attitudes: Controlling LLMs for Diverse Data Annotation* (No. arXiv:2410.11745). arXiv. http://arxiv.org/abs/2410.11745

Gamer, M., Lemon, J., & Singh, I. F. P. (2019). *irr: Various Coefficients of Interrater Reliability and Agreement*. https://CRAN.R-project.org/package=irr

Gentzkow, M., Kelly, B., & Taddy, M. (2019). Text as Data. *Journal of Economic Literature*, *57*(3), 535–574. https://doi.org/10.1257/jel.20181020

GESIS (2024). GESIS Panel - Extended Edition. *GESIS, Cologne. ZA5664 Data file Version 52.0.0.* https://doi.org/10.4232/1.14284

Grimmer, J., & Stewart, B. M. (2013). Text as Data: The Promise and Pitfalls of Automatic Content Analysis Methods for Political Texts. *Political Analysis*, *21*(3), 267–297. https://doi.org/10.1093/pan/mps028

Gweon, H., & Schonlau, M. (2024). Automated Classification for Open-Ended Questions with BERT. *Journal of Survey Statistics and Methodology*, *12*(2), 493–504.

https://doi.org/10.1093/jssam/smad015

Götz, F. M., Maertens, R., Loomba, S., & Van Der Linden, S. (2023). Let the algorithm speak: How to use neural networks for automatic item generation in psychological scale development. *Psychological Methods*. https://doi.org/10.1037/met0000540

Haensch, A.-C., Weiß, B., Steins, Patricia, Chyvra, Priscilla, & Bitz, Katja. (2022). The semi-automatic classification of an open-ended question on panel survey motivation and its application in attrition analysis. *Frontiers in Big Data*, *5:880554*. https://doi.org/10.3389/fdata.2022.880554

Halterman, A., & Keith, K. A. (2024). *Codebook LLMs: Adapting Political Science Codebooks for LLM Use and Adapting LLMs to Follow Codebooks* (No. arXiv:2407.10747). arXiv. http://arxiv.org/abs/2407.10747

Hand, D. J., Christen, P., & Ziyad, S. (2024). *Selecting a classification performance measure: Matching the measure to the problem* (No. arXiv:2409.12391). arXiv. http://arxiv.org/abs/2409.12391

Heseltine, M., & Clemm von Hohenberg, B. (2024). Large language models as a substitute for human experts in annotating political text. *Research & Politics*, *11*(1), 20531680241236239. https://doi.org/10.1177/20531680241236239

Joachims, T. (2001). A statistical learning model of text classification for support vector machines. *Proceedings of the 24th Annual International ACM SIGIR Conference on Research and Development in Information Retrieval*, 128–136. https://doi.org/10.1145/383952.383974

Johnson, R. L., Pistilli, G., Menédez-González, N., Duran, L. D. D., Panai, E., Kalpokiene, J., & Bertulfo, D. J. (2022). *The Ghost in the Machine has an American accent: Value conflict in GPT-3* (No. arXiv:2203.07785). arXiv. http://arxiv.org/abs/2203.07785

Kim, J., & Lee, B. (2023). *AI-Augmented Surveys: Leveraging Large Language Models and Surveys for Opinion Prediction* (No. arXiv:2305.09620). arXiv. http://arxiv.org/abs/2305.09620

Krosnick, J. A., & Presser, S. (2010). Question and Questionnaire Design. In P. V. Marsden & J. D. Wright (Eds.), *Handbook of Survey Research. 2nd edition* (pp. 263–314). Emerald.

Kuhn, M. (2008). Building Predictive Models in R Using the caret Package. *Journal of Statistical Software*, *28*(5), 1–26. https://doi.org/10.18637/jss.v028.i05

Landesvatter, C. (2024). *Methods for the classification of data from open-ended questions in surveys* [University of Mannheim]. https://madoc.bib.uni-mannheim.de/67089/

Lebrun, B., Temtsin, S., Vonasch, A., & Bartneck, C. (2024). Detecting the corruption of online questionnaires by artificial intelligence. *Frontiers in Robotics and AI*, *Volume 10-2023*. https://doi.org/10.3389/frobt.2023.1277635

Li, B., Haider, S., & Callison-Burch, C. (2024). This Land is Your, My Land: Evaluating Geopolitical Bias in Language Models through Territorial Disputes. *Proceedings of the 2024 Conference of the North American Chapter of the Association for Computational Linguistics: Human Language Technologies (Volume 1: Long Papers)*, 3855–3871. https://doi.org/10.18653/v1/2024.naacl-long.213

McCoy, R. T., Yao, S., Friedman, D., Hardy, M., & Griffiths, T. L. (2023). *Embers of Autoregression: Understanding Large Language Models Through the Problem They are Trained to Solve* (No. arXiv:2309.13638). arXiv. http://arxiv.org/abs/2309.13638

McKinney, W. (2010). Data Structures for Statistical Computing in Python. In S. van der Walt & J. Millman (Eds.), *Proceedings of the 9th Python in Science Conference* (pp. 56–61). https://doi.org/10.25080/Majora-92bf1922-00a

Meidinger, M., & Aßenmacher, M. (2021). A New Benchmark for NLP in Social Sciences: Evaluating the Usefulness of Pre-trained Language Models for Classifying Open-ended Survey Responses: *Proceedings of the 13th International Conference on Agents and Artificial Intelligence*, 866–873. https://doi.org/10.5220/0010255108660873

Mellon, J., Bailey, J., Scott, R., Breckwoldt, J., Miori, M., & Schmedeman, P. (2024). Do AIs know what the most important issue is? Using language models to code open-text social survey responses at scale. *Research & Politics*, *11*(1). https://doi.org/10.1177/20531680241231468

Meta. (2024a). *Llama 3.2: Revolutionizing edge AI and vision with open, customizable models*. Meta. https://ai.meta.com/blog/llama-3-2-connect-2024-vision-edge-mobile-devices/

Meta. (2024b). *Llama-3.2-3B-Instruct*. HuggingFace. https://huggingface.co/meta-llama/Llama-3.2-3B-Instruct

MistralAI. (2024a). *Mistral NeMo*. Mistral AI. https://mistral.ai/news/mistral-nemo/

MistralAI. (2024b). *Mistral-Nemo-Instruct-2407*. HuggingFace. https://huggingface.co/mistralai/Mistral-Nemo-Instruct-2407

Ooi, H., httr development team, Littlefield, T., Holden, S., Stone, C., & Microsoft. (2019). *AzureAuth: Authentication Services for Azure Active Directory* (p. 1.3.3) [Computer software]. 10.32614/CRAN.package.AzureAuth

OpenAI. (2024a). *Hello GPT-4o*. OpenAI. https://openai.com/index/hello-gpt-4o/

OpenAI. (2024b). *GPT-4o System Card*. https://cdn.openai.com/gpt-4o-system-card.pdf

Ornstein, J. T., Blasingame, E. N., & Truscott, J. S. (2024). *How to Train Your Stochastic Parrot: Large Language Models for Political Texts*. https://joeornstein.github.io/publications/ornstein-blasingame-truscott.pdf

Palmer, A., Smith, N. A., & Spirling, A. (2023). Using proprietary language models in academic research requires explicit justification. *Nature Computational Science*, *4*(1), 2–3. https://doi.org/10.1038/s43588-023-00585-1

Pangakis, N., Wolken, S., & Fasching, N. (2023). *Automated Annotation with Generative AI Requires Validation* (No. arXiv:2306.00176). arXiv. https://doi.org/10.48550/arXiv.2306.00176

Paszke, A., Gross, S., Massa, F., Lerer, A., Bradbury, J., Chanan, G., Killeen, T., Lin, Z., Gimelshein, N., Antiga, L., Desmaison, A., Kopf, A., Yang, E., DeVito, Z., Raison, M., Tejani, A., Chilamkurthy, S., Steiner, B., Fang, L., … Chintala, S. (2019). PyTorch: An Imperative Style, High-Performance Deep Learning Library. In *Advances in Neural Information Processing Systems 32* (pp. 8024–8035). Curran Associates, Inc. http://papers.neurips.cc/paper/9015-pytorch-an-imperative-style-high-performance-deep-learning-library.pdf

Pezeshkpour, P., & Hruschka, E. (2024). Large Language Models Sensitivity to The Order of Options in Multiple-Choice Questions. In K. Duh, H. Gomez, & S. Bethard (Eds.), *Findings of the Association for Computational Linguistics: NAACL 2024* (pp. 2006–2017). Association for Computational Linguistics. https://doi.org/10.18653/v1/2024.findings-naacl.130

Pham, C., Hoyle, A., Sun, S., Resnik, P., & Iyyer, M. (2024). TopicGPT: A Prompt-based Topic Modeling Framework. In K. Duh, H. Gomez, & S. Bethard (Eds.), *Proceedings of the 2024 Conference of the North American Chapter of the Association for Computational Linguistics: Human Language Technologies (Volume 1: Long Papers)* (pp. 2956–2984). Association for Computational Linguistics. https://doi.org/10.18653/v1/2024.naacl-long.164

Porst, R., & von Briel, C. (1995). *Wären Sie vielleicht bereit, sich gegebenenfalls noch einmal befragen zu lassen? Oder: Gründe für die Teilnahme an Panelbefragungen.* (No. 95/04; ZUMA Arbeitsbericht). Zentrum für Umfragen, Methoden und Analysen. https://www.gesis.org/fileadmin/upload/forschung/publikationen/gesis_reihen/zuma_arbeitsberichte/95_04.pdf

R Core Team. (2024, July 2). *R: The R Project for Statistical Computing*. https://www.r-project.org/

Rytting, C. M., Sorensen, T., Argyle, L., Busby, E., Fulda, N., Gubler, J., & Wingate, D. (2023). *Towards Coding Social Science Datasets with Language Models* (No. arXiv:2306.02177). arXiv. http://arxiv.org/abs/2306.02177

Schierholz, M., & Schonlau, M. (2021). Machine Learning for Occupation Coding—A Comparison Study. *Journal of Survey Statistics and Methodology*, *9*(5), 1013–1034. https://doi.org/10.1093/jssam/smaa023

Schonlau, M., & Couper, M. P. (2016). Semi-automated categorization of open-ended questions. *Survey Research Methods*, *Vol 10*, 143-152 Pages. https://doi.org/10.18148/SRM/2016.V10I2.6213

Schonlau, M., Weiß, J., & Marquardt, J. (2023). Multi-label classification of open-ended questions with BERT. *2023 Big Data Meets Survey Science (BigSurv)*, 1–8. https://doi.org/10.1109/BigSurv59479.2023.10486634

Schulhoff, S., Ilie, M., Balepur, N., Kahadze, K., Liu, A., Si, C., Li, Y., Gupta, A., Han, H., Schulhoff, S., Dulepet, P. S., Vidyadhara, S., Ki, D., Agrawal, S., Pham, C., Kroiz, G., Li, F., Tao, H., Srivastava, A., Da Costa, H., Gupta, S., Rogers, M. L., Goncearenco, I., Sarli, G., Galynker, I., Peskoff, D., Carpuat, M., White, J., Anadkat, S., Hoyle, A., & Resnik, P. (2024). *The prompt report: A systematic survey of prompting techniques.* (No. arXiv:2406.06608). arXiv. https://arxiv.org/abs/2406.06608

Spirling, A. (2023). Why open-source generative AI models are an ethical way forward for science. *Nature*, *616*(7957), 413–413. https://doi.org/10.1038/d41586-023-01295-4

Tian, K., Mitchell, E., Zhou, A., Sharma, A., Rafailov, R., Yao, H., Finn, C., & Manning, C. (2023). Just Ask for Calibration: Strategies for Eliciting Calibrated Confidence Scores from Language Models Fine-Tuned with Human Feedback. In H. Bouamor, J. Pino, & K. Bali (Eds.), *Proceedings of the 2023 Conference on Empirical Methods in Natural Language Processing* (pp. 5433–5442). Association for Computational Linguistics. https://doi.org/10.18653/v1/2023.emnlp-main.330

Törnberg, P. (2024). Large Language Models Outperform Expert Coders and Supervised Classifiers at Annotating Political Social Media Messages. *Social Science Computer Review*, 08944393241286471. https://doi.org/10.1177/08944393241286471

von der Heyde, L., Haensch, A.-C., Wenz, A., & Ma, B. (2025). *United in Diversity? Contextual Biases in LLM-Based Predictions of the 2024 European Parliament Elections* (No. arXiv: 2409.09045). arXiv. https://arxiv.org/abs/2409.09045

Wang, W., Jiao, W., Huang, J., Dai, R., Huang, J., Tu, Z., & Lyu, M. (2024). Not All Countries Celebrate Thanksgiving: On the Cultural Dominance in Large Language Models. In L.-W. Ku, A. Martins, & V. Srikumar (Eds.), *Proceedings of the 62nd Annual Meeting of the Association for Computational Linguistics (Volume 1: Long Papers)* (pp. 6349–6384). Association for Computational Linguistics. https://aclanthology.org/2024.acl-long.345

Weber, M., & Reichardt, M. (2023). *Evaluation is all you need. Prompting Generative Large Language Models for Annotation Tasks in the Social Sciences. A Primer using Open Models* (No. arXiv:2401.00284). arXiv. http://arxiv.org/abs/2401.00284

Wickham, H., Averick, M., Bryan, J., Chang, W., McGowan, L., François, R., Grolemund, G., Hayes, A., Henry, L., Hester, J., Kuhn, M., Pedersen, T., Miller, E., Bache, S., Müller, K., Ooms, J., Robinson, D., Seidel, D., Spinu, V., … Yutani, H. (2019). Welcome to the Tidyverse. *Journal of Open Source Software*, *4*(43), Article 43. https://doi.org/10.21105/joss.01686

Wolf, T., Debut, L., Sanh, V., Chaumond, J., Delangue, C., Moi, A., Cistac, P., Ma, C., Jernite, Y., Plu, J., Xu, C., Le Scao, T., Gugger, S., Drame, M., Lhoest, Q., & Rush, A. M. (2020). *Transformers: State-of-the-Art Natural Language Processing* (Version v4.25.1) [Computer software]. Zenodo. https://doi.org/10.5281/zenodo.7391177

Zhang, X., Li, S., Hauer, B., Shi, N., & Kondrak, G. (2023). Don't Trust ChatGPT when your Question is not in English: A Study of Multilingual Abilities and Types of LLMs. *Proceedings of the 2023 Conference on Empirical Methods in Natural Language Processing*, 7915–7927. https://doi.org/10.18653/v1/2023.emnlp-main.491

*Accelerate*. (n.d.). HuggingFace. Retrieved January 30, 2025, from https://huggingface.co/docs/accelerate/index

*Hub client library*. (n.d.). HuggingFace. Retrieved January 30, 2025, from https://huggingface.co/docs/huggingface_hub/index

**Appendix I:** (see online supplementary materials upon publication)
Coding scheme, descriptions and examples in German and English

**Appendix II**

  *1.  Survey Question*

**(2) Aus welchen Gründen nehmen Sie an den Umfragen des GESIS GesellschaftsMonitors teil?**
Bitte nennen Sie die drei wichtigsten Gründe.

Wichtigster Grund: _______________________________

Zweitwichtigster Grund: _______________________________

Drittwichtigster Grund: _______________________________

*Figure A1: Question on survey motivation as asked in the GESIS Panel.pop (Bosnjak et al., 2018, GESIS, 2024).*
*Translation: "(2) For what reasons do you participate in the surveys of the GESIS GesellschaftsMonitor? Please name the three most important reasons. Most important reason: … , Second most important reason: … Third most important reason: … "*

2. Prompting

*Du bist eine Expertin für Umfragen, die offene Antworten auf die Frage, wieso Personen an einer Umfrage teilnehmen, klassifiziert. Ordne diese Teilnahmegründe genau einer der folgenden Kategorien zu.*

*Die Kategorien sind:*

*[GERMAN CATEGORY 1: German description]*
*[GERMAN CATEGORY 2: German description]*
*[...]*
*[GERMAN CATEGORY 22: German description]*

*Stelle deine bestmögliche Vermutung an, auch wenn es schwer fällt.*
*Antworte im folgenden Format: Teilnahmegrund | KATEGORIE.*
*Begründe deine Zuordnung nicht, sondern gib nur den Teilnahmegrund und deine Zuordnung zurück.*

*Beispiele:*
*[German example reason | GERMAN CATEGORY 1]*
*[German example reason | GERMAN CATEGORY 2]*
*[...]*
*[German example reason | GERMAN CATEGORY 22]*

*Klassifiziere den folgenden Teilnahmegrund:*

*[German open-ended response]*

Figure A2: German prompt used for LLM-based classifications of the open-ended survey question. Categories and, in the detailed approach, descriptions (green font) were randomized across individual queries. In the few-shot approach, examples (blue font) were randomly selected, the selection being held constant, but presented in random order across queries. For details of descriptions and examples used, see Appendix I.

*3. Failed classifications*

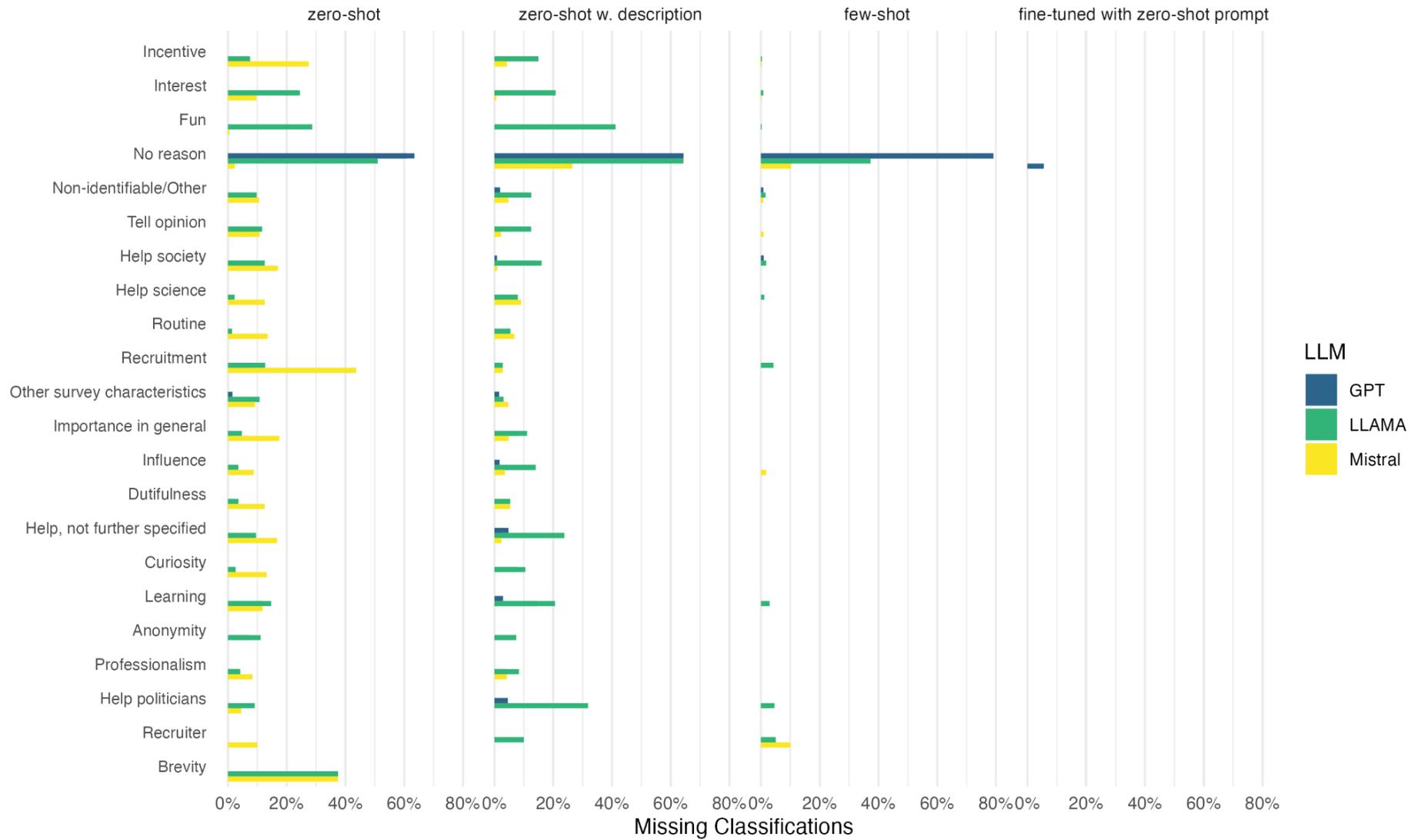

Figure A3: Proportion of missing classifications by category, LLM, and prompting approach.

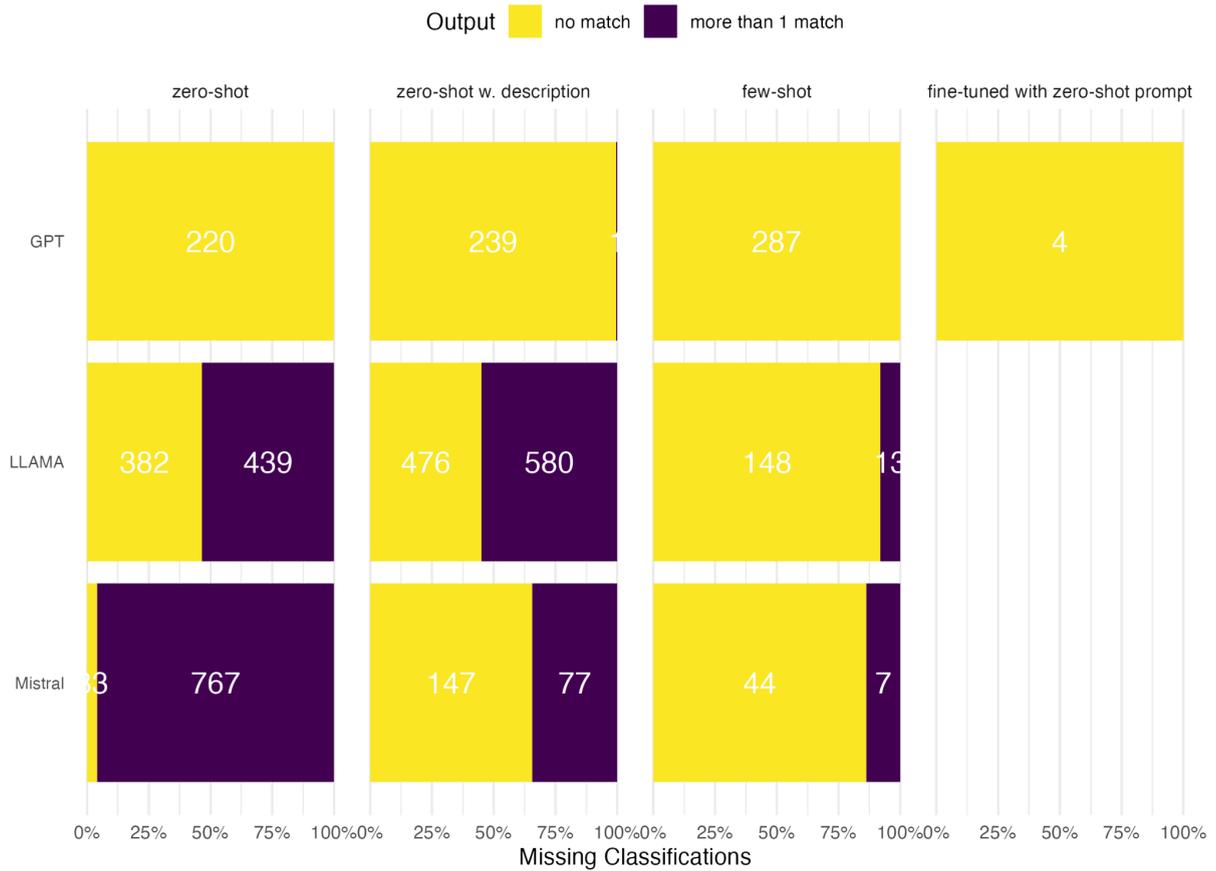

*Figure A4: Distribution of outputs recorded as missing due to missing or ambiguous classifications.*

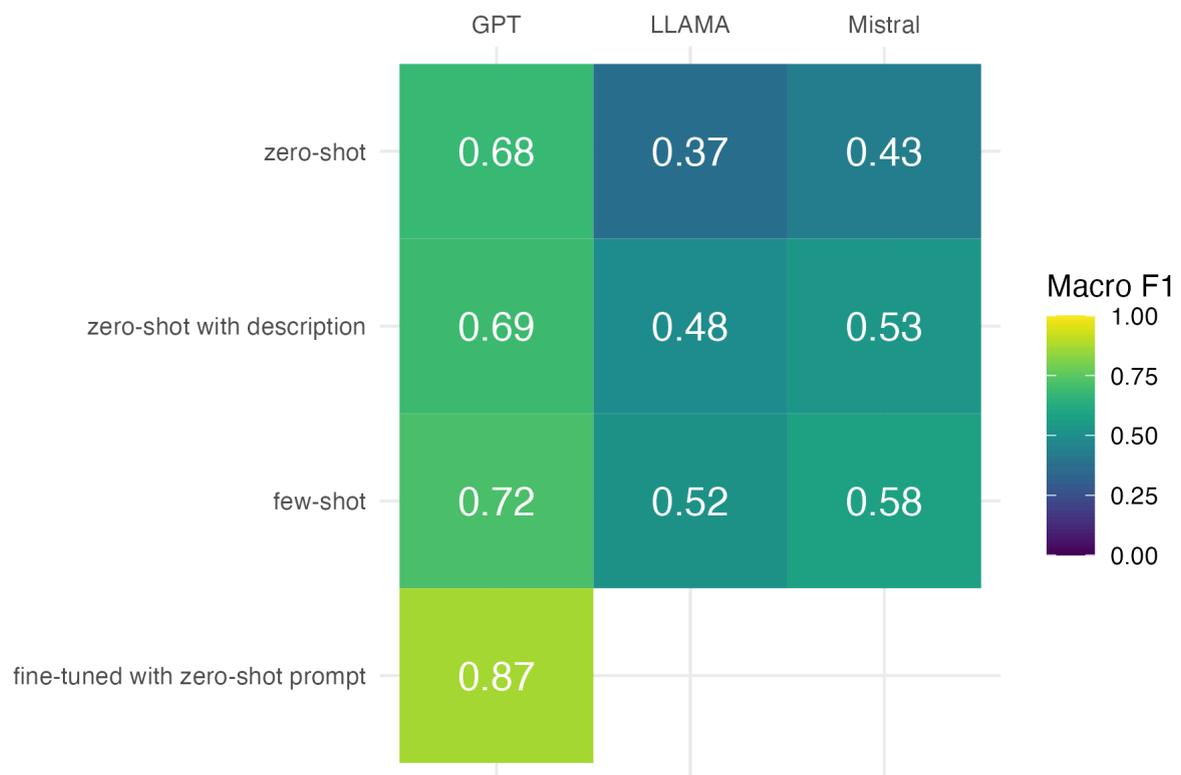

*Figure A5: Macro F1 scores by LLM and prompting approach without missing values.*

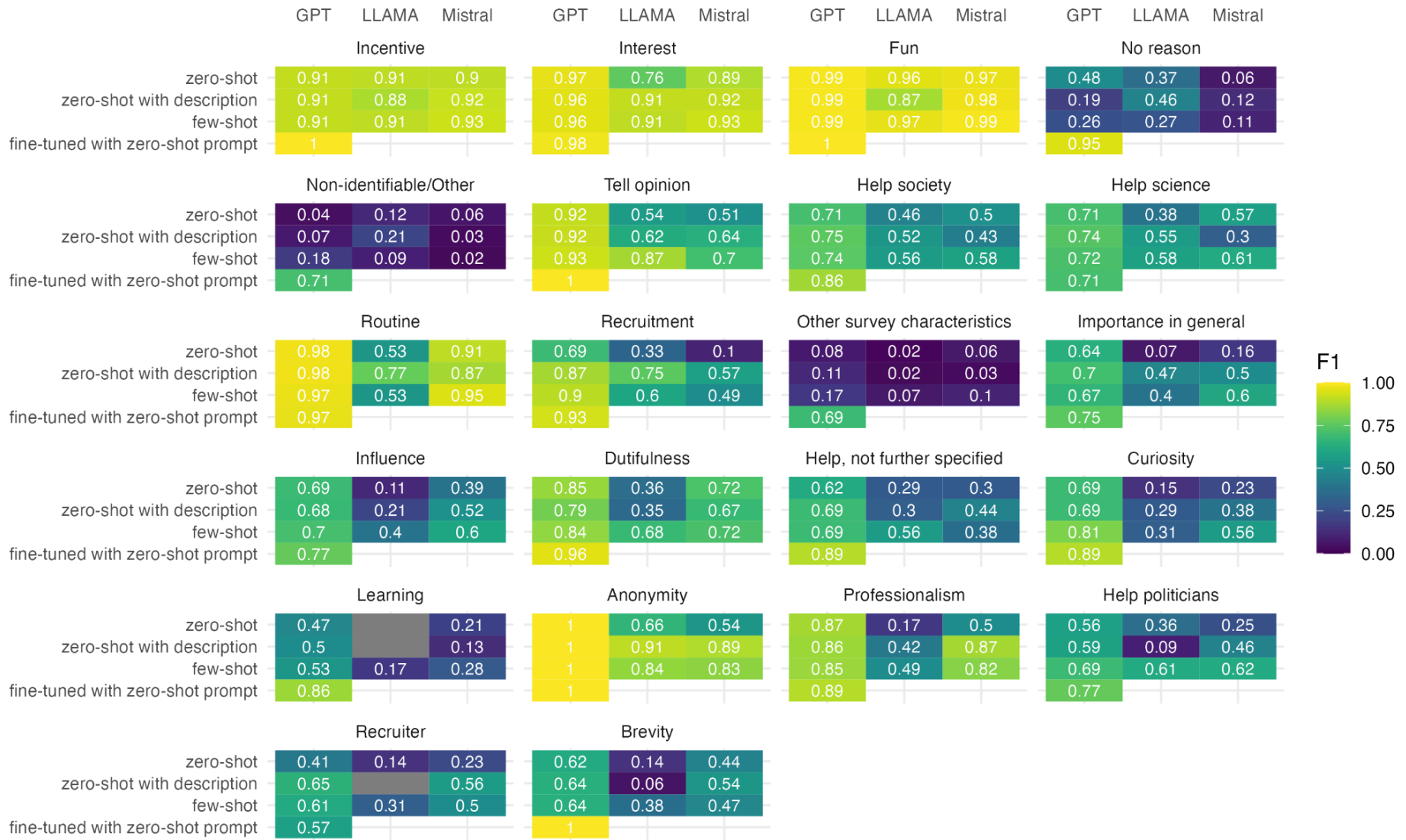

*Figure A6: Per-category scores by LLM and prompting approach without missing values.*

*4. Additional performance metrics*

**F1 Scores**

F1 scores evaluate machine learning model accuracy by considering model precision and recall. Recall refers to the number of correct classifications for a given category (true positives) divided by the number of actual classifications for that category (true positives + false negatives), measuring the share of classifications for a category as coded by the human experts that were correctly classified by the LLM. Precision refers to the number of correct classifications for a given category (true positives) divided by the number of all classifications for that category (true + false positives), measuring the share of LLM classifications for a given category that were correct. The F1 score is defined as 2*precision*recall / (precision + recall), with a range of [0;1].
  *Macro F1 scores* first calculate the F1 for each category separately and then average across these scores for a total model performance score, thereby taking into account unbalanced datasets, which may result in the LLM just assigning the modal category.
  *Weighted F1 scores* adjust for class imbalance by taking a weighted average of per-category F1 scores based on the number of true instances in each category. They are calculated as the weighted harmonic mean of precision and recall. This helps balance the contributions of minority categories without disproportionately emphasizing them. They thus offer a compromise between macro F1 and accuracy (see below), balancing performance on both common and rare classes without over-penalizing a model that performs well on the modal category.

**Accuracy**

Accuracy describes the proportion of cases correctly classified out of all cases.

**Intraclass Correlation Coefficient (ICC)**

The ICC quantifies the agreement raters by evaluating the proportion of variance attributable to differences between raters, relative to the total variance. The type used here is ICC(2,1), which assumes a two-way random-effects model and measures agreement. Thus, it does not consider the human-generated classifications as a benchmark or ground truth, but simply as a different coder to compare the LLM to.

**Cohen's Kappa**

Cohen's Kappa is a measure of agreement that takes into account the agreement occurring by chance, expressed as a proportion of the total possible improvement. It is particularly helpful for imbalanced datasets, as it quantifies model performance relative to a baseline of random chance, i.e., a naive classifier.

| Metric | Approach | GPT-4o | Llama-3.2 | Mistral NeMo |
|---|---|---|---|---|
| **Weighted F1** | *zero-shot* | 0.72 | 0.40 | 0.46 |
| | *with descriptions* | 0.72 | 0.42 | 0.63 |
| | *few-shot* | 0.72 | 0.63 | 0.67 |
| | *fine-tuned* | 0.91 | | |
| **Accuracy** | *zero-shot* | 0.81 | 0.55 | 0.60 |
| | *with descriptions* | 0.79 | 0.58 | 0.74 |
| | *few-shot* | 0.80 | 0.74 | 0.77 |
| | *fine-tuned* | 0.95 | | |
| **ICC** | *zero-shot* | 0.83 | 0.37 | 0.61 |
| | *with descriptions* | 0.81 | 0.47 | 0.67 |
| | *few-shot* | 0.84 | 0.65 | 0.74 |
| | *fine-tuned* | 0.96 | | |
| **Cohen's Kappa** | *zero-shot* | 0.76 | 0.47 | 0.53 |
| | *with descriptions* | 0.75 | 0.50 | 0.68 |
| | *few-shot* | 0.75 | 0.68 | 0.72 |
| | *fine-tuned* | 0.94 | | |

Table A1: Classification performance metrics per LLM and prompting approach, compared to human labels.

## 5. Confusion Matrices

**GPT** – zero-shot

| R: Prediction / C: Reference | Anonymity | Brevity | Curiosity | Dutifulness | Fun | Help politicians | Help science | Help society | Help, not further specified | Importance in general | Incentive | Influence | Interest | Learning | No reason | Non-identifiable/Other | Other survey characteristics | Professionalism | Recruiter | Recruitment | Routine | Tell opinion |
|---|---|---|---|---|---|---|---|---|---|---|---|---|---|---|---|---|---|---|---|---|---|---|
| **Anonymity** | 27 | 0 | 0 | 0 | 0 | 0 | 0 | 0 | 0 | 0 | 0 | 0 | 0 | 0 | 0 | 0 | 0 | 0 | 0 | 0 | 0 | 0 |
| **Brevity** | 0 | 8 | 0 | 0 | 0 | 0 | 0 | 0 | 0 | 0 | 0 | 0 | 0 | 0 | 0 | 0 | 10 | 0 | 0 | 0 | 0 | 0 |
| **Curiosity** | 0 | 0 | 38 | 0 | 0 | 0 | 0 | 0 | 0 | 0 | 0 | 0 | 10 | 3 | 0 | 20 | 1 | 0 | 0 | 0 | 0 | 0 |
| **Dutifulness** | 0 | 0 | 0 | 48 | 0 | 0 | 0 | 1 | 1 | 0 | 0 | 0 | 0 | 0 | 0 | 6 | 0 | 0 | 0 | 0 | 1 | 0 |
| **Fun** | 0 | 0 | 0 | 0 | 376 | 0 | 0 | 0 | 0 | 0 | 0 | 0 | 1 | 1 | 0 | 2 | 5 | 0 | 0 | 0 | 0 | 0 |
| **Help politicians** | 0 | 0 | 0 | 0 | 0 | 22 | 1 | 1 | 1 | 0 | 0 | 7 | 2 | 1 | 0 | 21 | 0 | 0 | 0 | 0 | 0 | 0 |
| **Help science** | 0 | 0 | 0 | 0 | 0 | 0 | 79 | 2 | 11 | 12 | 0 | 3 | 1 | 0 | 0 | 26 | 0 | 1 | 0 | 0 | 0 | 0 |
| **Help society** | 0 | 0 | 0 | 0 | 0 | 0 | 0 | 96 | 7 | 5 | 0 | 9 | 1 | 1 | 0 | 40 | 0 | 0 | 0 | 0 | 0 | 0 |
| **Help, not further specified** | 0 | 0 | 0 | 0 | 0 | 0 | 0 | 0 | 20 | 0 | 0 | 0 | 0 | 0 | 0 | 2 | 0 | 0 | 0 | 1 | 0 | 0 |
| **Importance in general** | 0 | 0 | 0 | 0 | 0 | 0 | 1 | 10 | 0 | 42 | 0 | 0 | 0 | 0 | 0 | 12 | 4 | 0 | 0 | 0 | 0 | 0 |
| **Incentive** | 0 | 0 | 0 | 5 | 0 | 0 | 0 | 0 | 0 | 0 | 1523 | 0 | 0 | 0 | 0 | 3 | 0 | 0 | 1 | 0 | 0 | 0 |
| **Influence** | 0 | 0 | 0 | 0 | 0 | 0 | 0 | 0 | 0 | 1 | 0 | 38 | 1 | 0 | 0 | 12 | 0 | 0 | 0 | 0 | 0 | 1 |
| **Interest** | 0 | 0 | 0 | 0 | 0 | 0 | 0 | 1 | 0 | 1 | 0 | 0 | 1234 | 6 | 0 | 27 | 31 | 0 | 1 | 0 | 0 | 0 |
| **Learning** | 0 | 0 | 0 | 0 | 0 | 0 | 4 | 0 | 1 | 2 | 0 | 0 | 0 | 21 | 0 | 28 | 0 | 0 | 0 | 0 | 0 | 0 |
| **No reason** | 0 | 0 | 0 | 1 | 0 | 0 | 0 | 0 | 0 | 0 | 270 | 0 | 2 | 0 | 134 | 4 | 2 | 0 | 1 | 3 | 0 | 0 |
| Non-identifiable/Other | 0 | 0 | 0 | 1 | 0 | 0 | 0 | 0 | 0 | 0 | 3 | 0 | 0 | 0 | 6 | 5 | 1 | 0 | 1 | 0 | 0 | 0 |
| Other survey characteristics | 0 | 0 | 0 | 0 | 0 | 0 | 1 | 0 | 1 | 0 | 1 | 0 | 0 | 0 | 0 | 9 | 3 | 0 | 0 | 0 | 0 | 0 |
| Professionalism | 0 | 0 | 0 | 0 | 0 | 0 | 0 | 0 | 0 | 0 | 0 | 0 | 0 | 0 | 0 | 0 | 0 | 6 | 23 | 0 | 0 | 0 | 0 |
| Recruiter | 0 | 0 | 0 | 0 | 0 | 0 | 0 | 0 | 0 | 0 | 0 | 0 | 0 | 0 | 0 | 0 | 0 | 0 | 0 | 12 | 26 | 0 | 0 |
| Recruitment | 0 | 0 | 0 | 1 | 0 | 0 | 0 | 0 | 0 | 0 | 0 | 0 | 0 | 0 | 0 | 1 | 0 | 0 | 4 | 41 | 1 | 0 |
| Routine | 0 | 0 | 0 | 0 | 0 | 0 | 0 | 0 | 0 | 0 | 0 | 0 | 0 | 0 | 0 | 1 | 0 | 0 | 0 | 0 | 72 | 0 |
| Tell opinion | 0 | 0 | 0 | 0 | 0 | 0 | 2 | 1 | 0 | 0 | 0 | 0 | 0 | 1 | 0 | 36 | 1 | 0 | 0 | 0 | 0 | 231 |
| NA | 0 | 0 | 0 | 0 | 0 | 0 | 0 | 0 | 0 | 0 | 0 | 0 | 0 | 0 | 219 | 0 | 1 | 0 | 0 | 0 | 0 | 0 |

*Table A2: Confusion matrix (actual vs. predicted categories) for GPT under zero-shot prompting.*

## GPT – with descriptions

| R: Prediction / C: Reference | Anonymity | Brevity | Curiosity | Dutifulness | Fun | Help politicians | Help science | Help society | Help, not further specified | Importance in general | Incentive | Influence | Interest | Learning | No reason | Non-identifiable/Other | Other survey characteristics | Professionalism | Recruiter | Recruitment | Routine | Tell opinion |
|---|---|---|---|---|---|---|---|---|---|---|---|---|---|---|---|---|---|---|---|---|---|---|
| **Anonymity** | 27 | 0 | 0 | 0 | 0 | 0 | 0 | 0 | 0 | 0 | 0 | 0 | 0 | 0 | 0 | 0 | 0 | 0 | 0 | 0 | 0 | 0 |
| **Brevity** | 0 | 8 | 0 | 0 | 0 | 0 | 0 | 0 | 0 | 0 | 0 | 0 | 0 | 0 | 0 | 0 | 9 | 0 | 0 | 0 | 0 | 0 |
| **Curiosity** | 0 | 0 | 38 | 0 | 0 | 0 | 1 | 0 | 0 | 1 | 0 | 0 | 12 | 1 | 0 | 18 | 1 | 0 | 0 | 0 | 0 | 0 |
| **Dutifulness** | 0 | 0 | 0 | 38 | 0 | 0 | 0 | 0 | 0 | 0 | 0 | 0 | 0 | 0 | 0 | 2 | 0 | 0 | 0 | 0 | 0 | 0 |
| **Fun** | 0 | 0 | 0 | 0 | 376 | 0 | 0 | 0 | 0 | 0 | 0 | 0 | 0 | 1 | 0 | 2 | 4 | 0 | 0 | 0 | 0 | 0 |
| **Help politicians** | 0 | 0 | 0 | 0 | 0 | 17 | 1 | 1 | 0 | 0 | 2 | 0 | 0 | 0 | 0 | 15 | 1 | 0 | 0 | 0 | 0 | 0 |
| **Help science** | 0 | 0 | 0 | 0 | 0 | 0 | 72 | 0 | 5 | 7 | 0 | 2 | 0 | 0 | 0 | 20 | 0 | 0 | 0 | 0 | 0 | 0 |
| **Help society** | 0 | 0 | 0 | 0 | 0 | 1 | 0 | 104 | 7 | 4 | 0 | 8 | 1 | 1 | 0 | 39 | 0 | 0 | 0 | 0 | 0 | 0 |
| **Help, not further specified** | 0 | 0 | 0 | 0 | 0 | 0 | 0 | 0 | 24 | 0 | 0 | 3 | 0 | 0 | 0 | 3 | 0 | 0 | 0 | 0 | 0 | 0 |
| **Importance in general** | 0 | 0 | 0 | 0 | 0 | 0 | 2 | 3 | 0 | 47 | 0 | 1 | 0 | 0 | 0 | 16 | 2 | 0 | 0 | 0 | 0 | 0 |
| **Incentive** | 0 | 0 | 0 | 6 | 0 | 0 | 0 | 0 | 0 | 0 | 1522 | 0 | 0 | 0 | 0 | 1 | 0 | 0 | 1 | 0 | 0 | 0 |
| **Influence** | 0 | 0 | 0 | 0 | 0 | 3 | 0 | 2 | 0 | 0 | 0 | 40 | 1 | 0 | 0 | 12 | 0 | 0 | 0 | 0 | 0 | 4 |
| **Interest** | 0 | 0 | 0 | 0 | 0 | 0 | 4 | 1 | 0 | 1 | 0 | 0 | 1235 | 9 | 0 | 40 | 35 | 0 | 0 | 0 | 0 | 0 |
| **Learning** | 0 | 0 | 0 | 0 | 0 | 0 | 3 | 0 | 0 | 2 | 0 | 0 | 0 | 22 | 0 | 27 | 0 | 0 | 0 | 0 | 0 | 0 |
| **No reason** | 0 | 0 | 0 | 0 | 0 | 0 | 0 | 0 | 0 | 0 | 204 | 0 | 0 | 0 | 36 | 3 | 2 | 0 | 0 | 0 | 0 | 0 |

| | | | | | | | | | | | | | | | | | | | | | |
|---|---|---|---|---|---|---|---|---|---|---|---|---|---|---|---|---|---|---|---|---|---|
| **Non-identifiable/Other** | 0 | 0 | 0 | 1 | 0 | 0 | 0 | 0 | 0 | 0 | 71 | 0 | 2 | 0 | 91 | 16 | 2 | 0 | 0 | 0 | 0 | 0 |
| **Other survey characteristics** | 0 | 0 | 0 | 0 | 0 | 0 | 1 | 0 | 1 | 0 | 0 | 0 | 0 | 0 | 0 | 3 | 4 | 0 | 0 | 0 | 0 | 0 |
| **Professionalism** | 0 | 0 | 0 | 0 | 0 | 0 | 2 | 0 | 2 | 0 | 0 | 0 | 0 | 0 | 0 | 0 | 4 | 24 | 0 | 0 | 0 | 0 |
| **Recruiter** | 0 | 0 | 0 | 1 | 0 | 0 | 0 | 0 | 0 | 0 | 0 | 0 | 1 | 0 | 0 | 0 | 0 | 0 | 12 | 3 | 0 | 0 |
| **Recruitment** | 0 | 0 | 0 | 7 | 0 | 0 | 0 | 0 | 0 | 0 | 0 | 0 | 0 | 0 | 1 | 2 | 0 | 0 | 7 | 68 | 0 | 0 |
| **Routine** | 0 | 0 | 0 | 3 | 0 | 0 | 0 | 0 | 0 | 0 | 0 | 0 | 0 | 0 | 0 | 0 | 0 | 0 | 0 | 0 | 74 | 0 |
| **Tell opinion** | 0 | 0 | 0 | 0 | 0 | 0 | 2 | 0 | 1 | 1 | 0 | 0 | 0 | 0 | 0 | 33 | 0 | 0 | 0 | 0 | 0 | 228 |
| **NA** | 0 | 0 | 0 | 0 | 0 | 1 | 0 | 1 | 2 | 0 | 0 | 1 | 0 | 0 | 231 | 3 | 1 | 0 | 0 | 0 | 0 | 0 |

*Table A3: Confusion matrix (actual vs. predicted categories) for GPT under zero-shot prompting with descriptions.*

## GPT – few-shot

| R: Prediction / C: Reference | Anonymity | Brevity | Curiosity | Dutifulness | Fun | Help politicians | Help science | Help society | Help, not further specified | Importance in general | Incentive | Influence | Interest | Learning | No reason | Non-identifiable/Other | Other survey characteristics | Professionalism | Recruiter | Recruitment | Routine | Tell opinion |
|---|---|---|---|---|---|---|---|---|---|---|---|---|---|---|---|---|---|---|---|---|---|---|
| **Anonymity** | 27 | 0 | 0 | 0 | 0 | 0 | 0 | 0 | 0 | 0 | 0 | 0 | 0 | 0 | 0 | 0 | 0 | 0 | 0 | 0 | 0 | 0 |
| **Brevity** | 0 | 8 | 0 | 0 | 0 | 0 | 0 | 0 | 0 | 0 | 0 | 0 | 0 | 0 | 0 | 9 | 0 | 0 | 0 | 0 | 0 | 0 |
| **Curiosity** | 0 | 0 | 38 | 0 | 0 | 0 | 0 | 0 | 0 | 0 | 0 | 0 | 5 | 2 | 2 | 9 | 0 | 0 | 0 | 0 | 0 | 0 |
| **Dutifulness** | 0 | 0 | 0 | 46 | 0 | 0 | 0 | 1 | 1 | 0 | 0 | 0 | 0 | 0 | 0 | 4 | 0 | 0 | 0 | 0 | 2 | 0 |
| **Fun** | 0 | 0 | 0 | 0 | 376 | 0 | 0 | 0 | 0 | 0 | 0 | 0 | 1 | 0 | 0 | 2 | 2 | 0 | 0 | 0 | 0 | 0 |
| **Help politicians** | 0 | 0 | 0 | 0 | 0 | 21 | 1 | 2 | 0 | 0 | 0 | 4 | 0 | 0 | 0 | 11 | 0 | 0 | 0 | 0 | 0 | 0 |
| **Help science** | 0 | 0 | 0 | 0 | 0 | 0 | 81 | 1 | 9 | 11 | 0 | 2 | 1 | 0 | 0 | 32 | 0 | 1 | 0 | 0 | 0 | 0 |
| **Help society** | 0 | 0 | 0 | 0 | 0 | 0 | 0 | 99 | 8 | 4 | 0 | 7 | 1 | 0 | 0 | 36 | 0 | 0 | 0 | 0 | 0 | 0 |
| **Help, not further specified** | 0 | 0 | 0 | 0 | 0 | 0 | 0 | 0 | 24 | 0 | 0 | 1 | 0 | 0 | 0 | 3 | 0 | 0 | 0 | 0 | 0 | 0 |
| **Importance in general** | 0 | 0 | 0 | 0 | 0 | 0 | 2 | 3 | 0 | 44 | 0 | 0 | 0 | 0 | 0 | 15 | 5 | 0 | 0 | 0 | 0 | 0 |
| **Incentive** | 0 | 0 | 0 | 5 | 0 | 0 | 0 | 0 | 0 | 0 | 1523 | 0 | 0 | 0 | 0 | 3 | 0 | 0 | 1 | 0 | 0 | 0 |
| **Influence** | 0 | 0 | 0 | 0 | 0 | 1 | 0 | 1 | 0 | 1 | 0 | 43 | 1 | 0 | 1 | 15 | 0 | 0 | 0 | 0 | 0 | 2 |
| **Interest** | 0 | 0 | 0 | 0 | 0 | 0 | 0 | 1 | 0 | 0 | 0 | 0 | 1193 | 3 | 0 | 21 | 27 | 0 | 0 | 0 | 0 | 0 |
| **Learning** | 0 | 0 | 0 | 0 | 0 | 0 | 3 | 2 | 0 | 3 | 0 | 0 | 1 | 28 | 0 | 33 | 0 | 0 | 0 | 0 | 0 | 1 |
| **No reason** | 0 | 0 | 0 | 0 | 0 | 0 | 0 | 0 | 0 | 0 | 250 | 0 | 2 | 0 | 50 | 3 | 4 | 0 | 0 | 0 | 0 | 0 |
| **Non-identifiable/Other** | 0 | 0 | 0 | 1 | 0 | 0 | 0 | 0 | 0 | 0 | 24 | 0 | 0 | 1 | 20 | 29 | 1 | 0 | 0 | 0 | 0 | 0 |

| | | | | | | | | | | | | | | | | | | | | | |
|---|---|---|---|---|---|---|---|---|---|---|---|---|---|---|---|---|---|---|---|---|---|
| **Other survey characteristics** | 0 | 0 | 0 | 0 | 0 | 0 | 0 | 0 | 0 | 0 | 0 | 0 | 46 | 0 | 0 | 7 | 11 | 0 | 0 | 0 | 0 | 0 |
| **Professionalism** | 0 | 0 | 0 | 0 | 0 | 0 | 1 | 0 | 0 | 0 | 0 | 0 | 0 | 0 | 0 | 6 | 23 | 0 | 0 | 0 | 0 |
| **Recruiter** | 0 | 0 | 0 | 1 | 0 | 0 | 0 | 0 | 0 | 0 | 0 | 0 | 1 | 0 | 0 | 0 | 0 | 0 | 11 | 3 | 0 | 0 |
| **Recruitment** | 0 | 0 | 0 | 3 | 0 | 0 | 0 | 0 | 0 | 0 | 0 | 0 | 0 | 0 | 0 | 0 | 0 | 0 | 8 | 68 | 1 | 0 |
| **Routine** | 0 | 0 | 0 | 0 | 0 | 0 | 0 | 0 | 0 | 0 | 0 | 0 | 0 | 0 | 0 | 1 | 0 | 0 | 0 | 0 | 71 | 0 |
| **Tell opinion** | 0 | 0 | 0 | 0 | 0 | 0 | 0 | 1 | 0 | 0 | 0 | 0 | 0 | 0 | 2 | 29 | 0 | 0 | 0 | 0 | 0 | 229 |
| **NA** | 0 | 0 | 0 | 0 | 0 | 0 | 0 | 1 | 0 | 0 | 0 | 0 | 0 | 0 | 284 | 2 | 0 | 0 | 0 | 0 | 0 | 0 |

*Table A4: Confusion matrix (actual vs. predicted categories) for GPT under few-shot prompting.*

**GPT** – fine-tuned

| R: Prediction / C: Reference | Anonymity | Brevity | Curiosity | Dutifulness | Fun | Help politicians | Help science | Help society | Help, not further specified | Importance in general | Incentive | Influence | Interest | Learning | No reason | Non-identifiable/Other | Other survey characteristics | Professionalism | Recruiter | Recruitment | Routine | Tell opinion |
|---|---|---|---|---|---|---|---|---|---|---|---|---|---|---|---|---|---|---|---|---|---|---|
| **Anonymity** | 6 | 0 | 0 | 0 | 0 | 0 | 0 | 0 | 0 | 0 | 0 | 0 | 0 | 0 | 0 | 0 | 0 | 0 | 0 | 0 | 0 | 0 |
| **Brevity** | 0 | 2 | 0 | 0 | 0 | 0 | 0 | 0 | 0 | 0 | 0 | 0 | 0 | 0 | 0 | 0 | 0 | 0 | 0 | 0 | 0 | 0 |
| **Curiosity** | 0 | 0 | 8 | 0 | 0 | 0 | 0 | 0 | 0 | 0 | 0 | 0 | 1 | 0 | 0 | 1 | 0 | 0 | 0 | 0 | 0 | 0 |
| **Dutifulness** | 0 | 0 | 0 | 11 | 0 | 0 | 0 | 0 | 0 | 0 | 0 | 0 | 0 | 0 | 0 | 0 | 0 | 0 | 0 | 0 | 0 | 0 |
| **Fun** | 0 | 0 | 0 | 0 | 76 | 0 | 0 | 0 | 0 | 0 | 0 | 0 | 0 | 0 | 0 | 0 | 0 | 0 | 0 | 0 | 0 | 0 |
| **Help politicians** | 0 | 0 | 0 | 0 | 0 | 5 | 0 | 0 | 0 | 0 | 0 | 0 | 0 | 0 | 0 | 3 | 0 | 0 | 0 | 0 | 0 | 0 |
| **Help science** | 0 | 0 | 0 | 0 | 0 | 0 | 11 | 0 | 0 | 0 | 0 | 0 | 1 | 0 | 0 | 1 | 0 | 0 | 0 | 0 | 0 | 0 |
| **Help society** | 0 | 0 | 0 | 0 | 0 | 0 | 0 | 19 | 0 | 1 | 0 | 1 | 0 | 0 | 0 | 0 | 0 | 0 | 0 | 0 | 0 | 0 |
| **Help, not further specified** | 0 | 0 | 0 | 0 | 0 | 0 | 0 | 0 | 8 | 0 | 0 | 1 | 0 | 0 | 0 | 0 | 0 | 0 | 0 | 0 | 0 | 0 |
| **Importance in general** | 0 | 0 | 0 | 0 | 0 | 0 | 4 | 1 | 0 | 12 | 0 | 0 | 0 | 0 | 0 | 0 | 2 | 0 | 0 | 0 | 0 | 0 |
| **Incentive** | 0 | 0 | 0 | 1 | 0 | 0 | 0 | 0 | 0 | 0 | 360 | 0 | 0 | 0 | 0 | 0 | 0 | 0 | 0 | 0 | 0 | 0 |
| **Influence** | 0 | 0 | 0 | 0 | 0 | 0 | 0 | 1 | 0 | 0 | 0 | 10 | 0 | 0 | 0 | 3 | 0 | 0 | 0 | 0 | 0 | 0 |
| **Interest** | 0 | 0 | 0 | 0 | 0 | 0 | 1 | 1 | 0 | 0 | 0 | 0 | 248 | 0 | 1 | 1 | 0 | 0 | 1 | 0 | 0 | 0 |
| **Learning** | 0 | 0 | 0 | 0 | 0 | 0 | 0 | 0 | 0 | 0 | 0 | 0 | 0 | 6 | 0 | 1 | 0 | 0 | 0 | 0 | 0 | 0 |
| **No reason** | 0 | 0 | 0 | 0 | 0 | 0 | 0 | 0 | 0 | 0 | 0 | 0 | 0 | 0 | 63 | 1 | 0 | 0 | 0 | 0 | 0 | 0 |
| **Non-identifiable/Other** | 0 | 0 | 0 | 0 | 0 | 0 | 2 | 1 | 1 | 0 | 0 | 0 | 1 | 1 | 4 | 34 | 1 | 0 | 0 | 0 | 0 | 0 |

| | | | | | | | | | | | | | | | | | | | | | |
|---|---|---|---|---|---|---|---|---|---|---|---|---|---|---|---|---|---|---|---|---|---|
| **Other survey characteristics** | 0 | 0 | 0 | 0 | 0 | 0 | 0 | 0 | 0 | 0 | 0 | 0 | 0 | 0 | 0 | 5 | 10 | 1 | 0 | 0 | 0 | 0 |
| **Professionalism** | 0 | 0 | 0 | 0 | 0 | 0 | 0 | 0 | 0 | 0 | 0 | 0 | 0 | 0 | 0 | 0 | 0 | 4 | 0 | 0 | 0 | 0 |
| **Recruiter** | 0 | 0 | 0 | 0 | 0 | 0 | 0 | 0 | 0 | 0 | 0 | 0 | 0 | 0 | 0 | 0 | 0 | 0 | 2 | 1 | 0 | 0 |
| **Recruitment** | 0 | 0 | 0 | 0 | 0 | 0 | 0 | 0 | 0 | 0 | 0 | 0 | 0 | 0 | 0 | 0 | 0 | 0 | 1 | 14 | 0 | 0 |
| **Routine** | 0 | 0 | 0 | 0 | 0 | 0 | 0 | 0 | 0 | 0 | 0 | 0 | 0 | 0 | 1 | 0 | 0 | 0 | 0 | 0 | 15 | 0 |
| **Tell opinion** | 0 | 0 | 0 | 0 | 0 | 0 | 0 | 0 | 0 | 0 | 0 | 0 | 0 | 0 | 0 | 0 | 0 | 0 | 0 | 0 | 0 | 47 |
| **NA** | 0 | 0 | 0 | 0 | 0 | 0 | 0 | 0 | 0 | 0 | 0 | 0 | 0 | 4 | 0 | 0 | 0 | 0 | 0 | 0 | 0 | 0 |

*Table A5: Confusion matrix (actual vs. predicted categories) for fine-tuned GPT with zero-shot prompt.*

**Llama** – zero-shot

| R: Prediction / C: Reference | Anonymity | Brevity | Curiosity | Dutifulness | Fun | Help politicians | Help science | Help society | Help, not further specified | Importance in general | Incentive | Influence | Interest | Learning | No reason | Non-identifiable/Other | Other survey characteristics | Professionalism | Recruiter | Recruitment | Routine | Tell opinion |
|---|---|---|---|---|---|---|---|---|---|---|---|---|---|---|---|---|---|---|---|---|---|---|
| **Anonymity** | 25 | 0 | 0 | 0 | 0 | 0 | 0 | 0 | 0 | 0 | 0 | 0 | 0 | 6 | 15 | 3 | 1 | 0 | 0 | 0 | 0 | 1 |
| **Brevity** | 0 | 3 | 0 | 1 | 0 | 0 | 0 | 0 | 0 | 0 | 15 | 0 | 0 | 0 | 7 | 2 | 3 | 0 | 0 | 4 | 2 | 0 |
| **Curiosity** | 0 | 0 | 34 | 0 | 0 | 2 | 2 | 1 | 0 | 2 | 26 | 2 | 287 | 8 | 23 | 18 | 6 | 0 | 1 | 0 | 0 | 7 |
| **Dutifulness** | 0 | 0 | 0 | 22 | 0 | 1 | 0 | 3 | 1 | 1 | 0 | 6 | 0 | 1 | 0 | 12 | 1 | 1 | 0 | 1 | 0 | 17 |
| **Fun** | 0 | 0 | 0 | 1 | 262 | 0 | 0 | 0 | 0 | 1 | 7 | 0 | 2 | 0 | 0 | 4 | 0 | 0 | 1 | 1 | 0 | 0 |
| **Help politicians** | 0 | 0 | 0 | 0 | 0 | 7 | 0 | 4 | 0 | 0 | 0 | 2 | 1 | 0 | 0 | 4 | 0 | 0 | 0 | 0 | 0 | 1 |
| **Help science** | 0 | 0 | 0 | 0 | 0 | 2 | 76 | 27 | 15 | 35 | 3 | 18 | 15 | 10 | 0 | 84 | 7 | 8 | 0 | 0 | 0 | 14 |
| **Help society** | 0 | 0 | 0 | 0 | 0 | 2 | 0 | 38 | 5 | 0 | 4 | 2 | 1 | 0 | 1 | 11 | 1 | 0 | 0 | 0 | 0 | 3 |
| **Help, not further specified** | 0 | 0 | 0 | 0 | 0 | 0 | 0 | 4 | 10 | 2 | 3 | 5 | 0 | 0 | 0 | 4 | 0 | 1 | 0 | 0 | 0 | 1 |
| **Importance in general** | 0 | 0 | 0 | 0 | 0 | 0 | 0 | 5 | 0 | 3 | 0 | 3 | 1 | 1 | 0 | 3 | 2 | 0 | 0 | 0 | 0 | 5 |
| **Incentive** | 0 | 0 | 0 | 2 | 0 | 0 | 0 | 0 | 0 | 0 | 1419 | 2 | 0 | 0 | 13 | 6 | 5 | 0 | 1 | 1 | 0 | 6 |
| **Influence** | 0 | 0 | 0 | 0 | 0 | 1 | 0 | 1 | 0 | 1 | 2 | 4 | 0 | 0 | 1 | 2 | 0 | 0 | 0 | 0 | 0 | 6 |
| **Interest** | 0 | 0 | 1 | 0 | 1 | 0 | 0 | 3 | 1 | 5 | 29 | 2 | 643 | 2 | 1 | 15 | 13 | 1 | 2 | 0 | 0 | 23 |
| **Learning** | 0 | 0 | 0 | 2 | 0 | 0 | 0 | 1 | 0 | 0 | 0 | 0 | 0 | 0 | 0 | 0 | 0 | 0 | 0 | 4 | 1 | 0 |
| **No reason** | 0 | 1 | 0 | 6 | 0 | 0 | 0 | 0 | 0 | 0 | 11 | 0 | 2 | 0 | 47 | 5 | 3 | 0 | 0 | 3 | 0 | 2 |
| **Non-identifiable/Other** | 0 | 0 | 0 | 1 | 1 | 0 | 0 | 0 | 1 | 1 | 13 | 0 | 3 | 1 | 32 | 18 | 2 | 0 | 2 | 0 | 0 | 3 |
| **Other survey characteristics** | 0 | 0 | 0 | 0 | 0 | 0 | 0 | 1 | 0 | 1 | 5 | 0 | 0 | 1 | 24 | 5 | 1 | 0 | 0 | 0 | 0 | 11 |
| **Professionalism** | 0 | 0 | 0 | 0 | 0 | 1 | 1 | 7 | 1 | 6 | 2 | 3 | 0 | 0 | 0 | 7 | 2 | 7 | 0 | 0 | 1 | 22 |
| **Recruiter** | 0 | 0 | 0 | 4 | 0 | 1 | 1 | 0 | 1 | 0 | 28 | 0 | 0 | 0 | 1 | 3 | 2 | 2 | 6 | 14 | 0 | 1 |
| **Recruitment** | 0 | 0 | 0 | 8 | 0 | 2 | 5 | 0 | 1 | 1 | 43 | 4 | 0 | 0 | 1 | 9 | 2 | 3 | 6 | 30 | 1 | 2 |
| **Routine** | 0 | 1 | 2 | 7 | 4 | 1 | 0 | 2 | 0 | 0 | 58 | 2 | 0 | 0 | 10 | 13 | 7 | 0 | 1 | 7 | 68 | 1 |
| **Tell opinion** | 0 | 0 | 0 | 0 | 0 | 0 | 1 | 1 | 2 | 1 | 0 | 0 | 0 | 0 | 0 | 3 | 0 | 0 | 0 | 0 | 0 | 80 |
| **NA** | 2 | 3 | 1 | 2 | 108 | 2 | 2 | 14 | 4 | 3 | 129 | 2 | 297 | 4 | 183 | 24 | 7 | 1 | 0 | 6 | 1 | 26 |

*Table A6: Confusion matrix (actual vs. predicted categories) for Llama under zero-shot prompting.*

**Llama** – with descriptions

| R: Prediction / C: Reference | Anonymity | Brevity | Curiosity | Dutifulness | Fun | Help politicians | Help science | Help society | Help, not further specified | Importance in general | Incentive | Influence | Interest | Learning | No reason | Non-identifiable/Other | Other survey characteristics | Professionalism | Recruiter | Recruitment | Routine | Tell opinion |
|---|---|---|---|---|---|---|---|---|---|---|---|---|---|---|---|---|---|---|---|---|---|---|
| **Anonymity** | 27 | 0 | 0 | 0 | 0 | 0 | 1 | 0 | 0 | 0 | 2 | 0 | 0 | 0 | 0 | 3 | 1 | 0 | 0 | 0 | 0 | 0 |
| **Brevity** | 0 | 7 | 0 | 0 | 0 | 0 | 0 | 0 | 0 | 0 | 0 | 1 | 0 | 0 | 0 | 1 | 9 | 0 | 0 | 0 | 0 | 0 |
| **Curiosity** | 0 | 0 | 38 | 0 | 0 | 0 | 5 | 0 | 0 | 2 | 21 | 1 | 39 | 5 | 24 | 20 | 2 | 0 | 0 | 0 | 0 | 4 |
| **Dutifulness** | 0 | 0 | 0 | 30 | 0 | 0 | 0 | 0 | 0 | 0 | 0 | 0 | 0 | 0 | 0 | 6 | 0 | 0 | 0 | 0 | 0 | 0 |
| **Fun** | 0 | 0 | 0 | 0 | 375 | 0 | 0 | 0 | 1 | 0 | 2 | 0 | 2 | 4 | 1 | 2 | 3 | 0 | 0 | 0 | 0 | 0 |
| **Help politicians** | 0 | 0 | 0 | 0 | 0 | 8 | 0 | 0 | 1 | 0 | 0 | 1 | 0 | 0 | 1 | 1 | 0 | 0 | 0 | 0 | 0 | 1 |
| **Help science** | 0 | 0 | 0 | 0 | 0 | 1 | 36 | 3 | 4 | 4 | 42 | 3 | 0 | 0 | 61 | 9 | 0 | 0 | 0 | 0 | 0 | 0 |
| **Help society** | 0 | 0 | 0 | 2 | 0 | 3 | 1 | 63 | 3 | 8 | 37 | 7 | 0 | 0 | 32 | 22 | 1 | 0 | 0 | 0 | 0 | 0 |
| **Help, not further specified** | 0 | 0 | 0 | 1 | 0 | 2 | 2 | 2 | 20 | 2 | 5 | 4 | 0 | 2 | 0 | 9 | 0 | 0 | 0 | 0 | 0 | 0 |
| **Importance in general** | 0 | 0 | 0 | 0 | 0 | 0 | 6 | 26 | 2 | 37 | 1 | 0 | 1 | 2 | 0 | 11 | 2 | 0 | 0 | 0 | 0 | 1 |
| **Incentive** | 0 | 0 | 0 | 6 | 0 | 1 | 0 | 0 | 0 | 0 | 1478 | 0 | 0 | 0 | 0 | 5 | 0 | 0 | 1 | 0 | 0 | 0 |
| **Influence** | 0 | 0 | 0 | 0 | 0 | 1 | 0 | 0 | 0 | 0 | 2 | 23 | 0 | 0 | 0 | 7 | 0 | 0 | 0 | 0 | 0 | 1 |
| **Interest** | 0 | 0 | 0 | 0 | 0 | 3 | 15 | 8 | 1 | 3 | 24 | 3 | 1201 | 13 | 10 | 62 | 28 | 0 | 1 | 0 | 0 | 6 |
| **Learning** | 0 | 0 | 0 | 0 | 0 | 0 | 8 | 1 | 0 | 1 | 3 | 1 | 0 | 4 | 3 | 7 | 0 | 0 | 0 | 0 | 0 | 1 |
| **No reason** | 0 | 1 | 0 | 0 | 0 | 0 | 0 | 2 | 1 | 0 | 6 | 2 | 0 | 0 | 19 | 6 | 1 | 0 | 0 | 0 | 0 | 4 |
| Non-identifiable/Other | 0 | 0 | 0 | 0 | 0 | 0 | 0 | 0 | 0 | 0 | 0 | 0 | 2 | 0 | 1 | 4 | 3 | 0 | 2 | 1 | 0 | 0 |
| Other survey characteristics | 0 | 0 | 0 | 0 | 0 | 0 | 1 | 0 | 0 | 0 | 0 | 0 | 0 | 0 | 0 | 2 | 1 | 0 | 0 | 0 | 0 | 0 |
| Professionalism | 0 | 0 | 0 | 0 | 0 | 0 | 1 | 1 | 1 | 0 | 0 | 0 | 0 | 0 | 0 | 1 | 3 | 23 | 0 | 0 | 0 | 0 |
| Recruiter | 0 | 0 | 0 | 2 | 0 | 0 | 0 | 0 | 0 | 0 | 1 | 0 | 0 | 0 | 0 | 2 | 0 | 0 | 10 | 1 | 0 | 0 |
| Recruitment | 0 | 0 | 0 | 10 | 1 | 0 | 0 | 2 | 5 | 0 | 29 | 5 | 0 | 1 | 17 | 17 | 4 | 0 | 6 | 67 | 3 | 0 |
| Routine | 0 | 0 | 0 | 2 | 0 | 0 | 0 | 0 | 0 | 0 | 5 | 0 | 0 | 1 | 7 | 1 | 1 | 0 | 0 | 0 | 66 | 0 |
| Tell opinion | 0 | 0 | 0 | 0 | 0 | 3 | 4 | 3 | 2 | 3 | 63 | 4 | 0 | 2 | 88 | 45 | 3 | 0 | 0 | 0 | 0 | 209 |
| NA | 0 | 0 | 0 | 3 | 0 | 0 | 8 | 1 | 1 | 3 | 76 | 2 | 7 | 0 | 95 | 12 | 3 | 1 | 0 | 2 | 5 | 5 |

*Table A7: Confusion matrix (actual vs. predicted categories) for Llama under zero-shot prompting with descriptions.*

## Llama – few-shot

| R: Prediction / C: Reference | Anonymity | Brevity | Curiosity | Dutifulness | Fun | Help politicians | Help science | Help society | Help, not further specified | Importance in general | Incentive | Influence | Interest | Learning | No reason | Non-identifiable/Other | Other survey characteristics | Professionalism | Recruiter | Recruitment | Routine | Tell opinion |
|---|---|---|---|---|---|---|---|---|---|---|---|---|---|---|---|---|---|---|---|---|---|---|
| **Anonymity** | 27 | 0 | 0 | 0 | 0 | 0 | 0 | 0 | 0 | 0 | 2 | 0 | 0 | 0 | 7 | 0 | 0 | 0 | 1 | 0 | 0 | 0 |
| **Brevity** | 0 | 7 | 0 | 0 | 0 | 0 | 0 | 0 | 0 | 0 | 9 | 0 | 0 | 0 | 3 | 0 | 8 | 0 | 0 | 2 | 0 | 0 |
| **Curiosity** | 0 | 0 | 38 | 0 | 1 | 0 | 0 | 0 | 0 | 0 | 58 | 2 | 16 | 15 | 59 | 11 | 3 | 0 | 0 | 0 | 0 | 2 |
| **Dutifulness** | 0 | 0 | 0 | 35 | 0 | 0 | 0 | 1 | 1 | 0 | 1 | 2 | 0 | 0 | 3 | 4 | 0 | 0 | 0 | 0 | 0 | 0 |
| **Fun** | 0 | 0 | 0 | 0 | 374 | 0 | 0 | 0 | 0 | 0 | 14 | 0 | 5 | 0 | 3 | 2 | 0 | 0 | 0 | 0 | 0 | 1 |
| **Help politicians** | 0 | 0 | 0 | 0 | 0 | 14 | 0 | 0 | 0 | 0 | 0 | 3 | 2 | 0 | 1 | 4 | 0 | 0 | 0 | 0 | 0 | 1 |
| **Help science** | 0 | 0 | 0 | 0 | 0 | 2 | 80 | 4 | 8 | 15 | 0 | 4 | 2 | 3 | 6 | 57 | 3 | 2 | 0 | 0 | 0 | 3 |
| **Help society** | 0 | 0 | 0 | 0 | 0 | 3 | 2 | 94 | 6 | 19 | 1 | 13 | 6 | 2 | 8 | 65 | 3 | 0 | 0 | 0 | 0 | 3 |
| **Help, not further specified** | 0 | 0 | 0 | 0 | 0 | 0 | 0 | 0 | 21 | 1 | 3 | 2 | 0 | 0 | 2 | 4 | 0 | 0 | 0 | 0 | 0 | 0 |
| **Importance in general** | 0 | 0 | 0 | 1 | 0 | 0 | 0 | 7 | 1 | 22 | 0 | 3 | 0 | 0 | 6 | 6 | 0 | 0 | 0 | 0 | 0 | 2 |
| **Incentive** | 0 | 0 | 0 | 4 | 0 | 0 | 1 | 0 | 0 | 0 | 1505 | 0 | 0 | 0 | 2 | 2 | 0 | 0 | 1 | 0 | 0 | 0 |
| **Influence** | 0 | 0 | 0 | 0 | 0 | 2 | 1 | 0 | 0 | 0 | 1 | 21 | 2 | 1 | 5 | 9 | 3 | 0 | 0 | 0 | 0 | 4 |
| **Interest** | 0 | 0 | 0 | 0 | 0 | 0 | 0 | 1 | 1 | 1 | 36 | 0 | 1107 | 1 | 3 | 10 | 19 | 0 | 0 | 1 | 0 | 10 |
| **Learning** | 0 | 0 | 0 | 0 | 0 | 0 | 2 | 0 | 0 | 0 | 0 | 0 | 0 | 4 | 3 | 4 | 0 | 1 | 0 | 0 | 0 | 1 |
| **No reason** | 0 | 1 | 0 | 2 | 0 | 0 | 0 | 0 | 0 | 1 | 36 | 0 | 0 | 0 | 43 | 6 | 3 | 0 | 0 | 0 | 0 | 0 |
| **Non-identifiable/Other** | 0 | 0 | 0 | 1 | 0 | 0 | 0 | 0 | 0 | 0 | 25 | 1 | 3 | 4 | 20 | 15 | 4 | 0 | 1 | 0 | 0 | 1 |
| **Other survey** | 0 | 0 | 0 | 0 | 0 | 0 | 0 | 0 | 1 | 0 | 3 | 1 | 100 | 1 | 17 | 15 | 7 | 0 | 0 | 0 | 0 | 0 |

| | | | | | | | | | | | | | | | | | | | | | |
|---|---|---|---|---|---|---|---|---|---|---|---|---|---|---|---|---|---|---|---|---|---|
| characteristics | | | | | | | | | | | | | | | | | | | | | |
| Professionalism | 0 | 0 | 0 | 1 | 0 | 0 | 0 | 1 | 1 | 4 | 2 | 1 | 0 | 2 | 3 | 11 | 5 | 19 | 1 | 0 | 0 | 2 |
| Recruiter | 0 | 0 | 0 | 3 | 0 | 0 | 1 | 0 | 0 | 0 | 2 | 0 | 0 | 0 | 5 | 3 | 0 | 2 | 12 | 30 | 0 | 0 |
| Recruitment | 0 | 0 | 0 | 3 | 0 | 0 | 0 | 0 | 0 | 0 | 1 | 0 | 0 | 0 | 4 | 1 | 0 | 0 | 2 | 34 | 0 | 1 |
| Routine | 0 | 0 | 0 | 6 | 0 | 0 | 0 | 2 | 0 | 0 | 93 | 0 | 0 | 0 | 17 | 7 | 6 | 0 | 1 | 1 | 74 | 0 |
| Tell opinion | 0 | 0 | 0 | 0 | 0 | 0 | 0 | 1 | 2 | 0 | 0 | 4 | 0 | 0 | 5 | 15 | 1 | 0 | 0 | 0 | 0 | 201 |
| NA | 0 | 0 | 0 | 0 | 1 | 1 | 1 | 1 | 0 | 0 | 5 | 0 | 9 | 1 | 134 | 4 | 0 | 0 | 1 | 3 | 0 | 0 |

*Table A8: Confusion matrix (actual vs. predicted categories) for Llama under few-shot prompting.*

**Mistral** – zero-shot

| R: Prediction / C: Reference | Anonymity | Brevity | Curiosity | Dutifulness | Fun | Help politicians | Help science | Help society | Help, not further specified | Importance in general | Incentive | Influence | Interest | Learning | No reason | Non-identifiable/Other | Other survey characteristics | Professionalism | Recruiter | Recruitment | Routine | Tell opinion |
|---|---|---|---|---|---|---|---|---|---|---|---|---|---|---|---|---|---|---|---|---|---|---|
| **Anonymity** | 27 | 0 | 0 | 2 | 0 | 0 | 0 | 4 | 1 | 2 | 21 | 1 | 1 | 1 | 1 | 8 | 3 | 0 | 1 | 0 | 0 | 0 |
| **Brevity** | 0 | 4 | 0 | 0 | 0 | 0 | 0 | 0 | 0 | 0 | 0 | 0 | 0 | 0 | 1 | 0 | 7 | 0 | 0 | 1 | 0 | 0 |
| **Curiosity** | 0 | 0 | 31 | 1 | 1 | 3 | 4 | 9 | 2 | 7 | 40 | 3 | 68 | 6 | 12 | 35 | 10 | 0 | 0 | 1 | 0 | 1 |
| **Dutifulness** | 0 | 0 | 0 | 31 | 0 | 0 | 0 | 0 | 0 | 0 | 0 | 1 | 0 | 0 | 1 | 2 | 1 | 1 | 0 | 0 | 0 | 0 |
| **Fun** | 0 | 0 | 0 | 0 | 372 | 0 | 0 | 0 | 0 | 0 | 7 | 0 | 3 | 1 | 0 | 4 | 5 | 0 | 1 | 0 | 0 | 0 |
| **Help politicians** | 0 | 0 | 0 | 0 | 0 | 5 | 0 | 0 | 0 | 0 | 0 | 0 | 1 | 0 | 10 | 3 | 0 | 0 | 0 | 0 | 0 | 0 |
| **Help science** | 0 | 0 | 0 | 0 | 0 | 3 | 52 | 4 | 5 | 6 | 2 | 0 | 3 | 1 | 1 | 26 | 0 | 2 | 0 | 0 | 0 | 1 |
| **Help society** | 0 | 0 | 0 | 2 | 0 | 1 | 0 | 46 | 4 | 6 | 3 | 4 | 0 | 0 | 1 | 23 | 0 | 0 | 0 | 0 | 0 | 0 |
| **Help, not further specified** | 0 | 0 | 0 | 1 | 0 | 2 | 0 | 1 | 9 | 0 | 1 | 6 | 0 | 1 | 0 | 2 | 1 | 0 | 1 | 1 | 0 | 0 |
| **Importance in general** | 0 | 0 | 0 | 0 | 0 | 1 | 1 | 9 | 3 | 8 | 5 | 4 | 0 | 4 | 0 | 6 | 3 | 0 | 0 | 0 | 0 | 1 |
| **Incentive** | 0 | 0 | 0 | 6 | 0 | 0 | 2 | 0 | 0 | 0 | 1087 | 1 | 0 | 0 | 0 | 5 | 0 | 0 | 1 | 0 | 0 | 0 |
| **Influence** | 0 | 0 | 0 | 1 | 0 | 2 | 0 | 1 | 0 | 1 | 5 | 19 | 0 | 0 | 3 | 11 | 0 | 0 | 0 | 0 | 0 | 2 |
| **Interest** | 0 | 0 | 2 | 1 | 1 | 2 | 7 | 5 | 3 | 11 | 74 | 1 | 1049 | 7 | 9 | 41 | 18 | 0 | 3 | 0 | 0 | 2 |
| **Learning** | 0 | 0 | 0 | 1 | 0 | 0 | 7 | 4 | 1 | 0 | 11 | 1 | 2 | 8 | 0 | 10 | 1 | 0 | 0 | 1 | 0 | 1 |
| **No reason** | 0 | 1 | 0 | 0 | 0 | 0 | 0 | 1 | 0 | 0 | 2 | 0 | 0 | 0 | 12 | 3 | 1 | 0 | 1 | 2 | 0 | 0 |
| **Non-identifiable/Other** | 0 | 0 | 0 | 0 | 0 | 0 | 0 | 0 | 0 | 0 | 0 | 0 | 1 | 0 | 2 | 7 | 0 | 0 | 1 | 0 | 0 | 0 |
| **Other survey characteristics** | 0 | 0 | 0 | 0 | 0 | 0 | 0 | 0 | 0 | 0 | 0 | 0 | 0 | 0 | 0 | 0 | 2 | 0 | 0 | 1 | 0 | 0 |

| | | | | | | | | | | | | | | | | | | | | | |
|---|---|---|---|---|---|---|---|---|---|---|---|---|---|---|---|---|---|---|---|---|---|
| **Professionalism** | 0 | 0 | 0 | 0 | 0 | 0 | 2 | 3 | 0 | 3 | 8 | 4 | 1 | 0 | 0 | 6 | 1 | 17 | 1 | 0 | 0 | 0 |
| **Recruiter** | 0 | 0 | 0 | 1 | 0 | 0 | 0 | 0 | 0 | 0 | 0 | 1 | 0 | 0 | 0 | 0 | 0 | 0 | 6 | 27 | 0 | 0 |
| **Recruitment** | 0 | 0 | 0 | 1 | 0 | 1 | 0 | 0 | 1 | 0 | 14 | 1 | 0 | 0 | 0 | 7 | 3 | 0 | 2 | 4 | 3 | 0 |
| **Routine** | 0 | 0 | 0 | 1 | 0 | 0 | 0 | 0 | 0 | 0 | 4 | 0 | 0 | 0 | 0 | 1 | 1 | 0 | 0 | 2 | 61 | 0 |
| **Tell opinion** | 0 | 0 | 0 | 0 | 0 | 1 | 2 | 6 | 6 | 8 | 20 | 5 | 2 | 1 | 298 | 28 | 2 | 2 | 0 | 0 | 0 | 199 |
| **NA** | 0 | 3 | 5 | 7 | 2 | 1 | 11 | 19 | 7 | 11 | 493 | 5 | 121 | 4 | 8 | 27 | 6 | 2 | 2 | 31 | 10 | 25 |

*Table A9: Confusion matrix (actual vs. predicted categories) for Mistral under zero-shot prompting.*

**Mistral** – with descriptions

| R: Prediction / C: Reference | Anonymity | Brevity | Curiosity | Dutifulness | Fun | Help politicians | Help science | Help society | Help, not further specified | Importance in general | Incentive | Influence | Interest | Learning | No reason | Non-identifiable/Other | Other survey characteristics | Professionalism | Recruiter | Recruitment | Routine | Tell opinion |
|---|---|---|---|---|---|---|---|---|---|---|---|---|---|---|---|---|---|---|---|---|---|---|
| **Anonymity** | 27 | 0 | 0 | 0 | 0 | 0 | 1 | 0 | 0 | 0 | 2 | 0 | 0 | 0 | 0 | 3 | 1 | 0 | 0 | 0 | 0 | 0 |
| **Brevity** | 0 | 7 | 0 | 0 | 0 | 0 | 0 | 0 | 0 | 0 | 0 | 1 | 0 | 0 | 0 | 1 | 9 | 0 | 0 | 0 | 0 | 0 |
| **Curiosity** | 0 | 0 | 38 | 0 | 0 | 0 | 5 | 0 | 0 | 2 | 21 | 1 | 39 | 5 | 24 | 20 | 2 | 0 | 0 | 0 | 0 | 4 |
| **Dutifulness** | 0 | 0 | 0 | 30 | 0 | 0 | 0 | 0 | 0 | 0 | 0 | 0 | 0 | 0 | 0 | 6 | 0 | 0 | 0 | 0 | 0 | 0 |
| **Fun** | 0 | 0 | 0 | 0 | 375 | 0 | 0 | 0 | 1 | 0 | 2 | 0 | 2 | 4 | 1 | 2 | 3 | 0 | 0 | 0 | 0 | 0 |
| **Help politicians** | 0 | 0 | 0 | 0 | 0 | 8 | 0 | 0 | 1 | 0 | 0 | 1 | 0 | 0 | 1 | 1 | 0 | 0 | 0 | 0 | 0 | 1 |
| **Help science** | 0 | 0 | 0 | 0 | 0 | 1 | 36 | 3 | 4 | 4 | 42 | 3 | 0 | 0 | 61 | 9 | 0 | 0 | 0 | 0 | 0 | 0 |
| **Help society** | 0 | 0 | 0 | 2 | 0 | 3 | 1 | 63 | 3 | 8 | 37 | 7 | 0 | 0 | 32 | 22 | 1 | 0 | 0 | 0 | 0 | 0 |
| **Help, not further specified** | 0 | 0 | 0 | 1 | 0 | 2 | 2 | 2 | 20 | 2 | 5 | 4 | 0 | 2 | 0 | 9 | 0 | 0 | 0 | 0 | 0 | 0 |
| **Importance in general** | 0 | 0 | 0 | 0 | 0 | 0 | 6 | 26 | 2 | 37 | 1 | 0 | 1 | 2 | 0 | 11 | 2 | 0 | 0 | 0 | 0 | 1 |
| **Incentive** | 0 | 0 | 0 | 6 | 0 | 1 | 0 | 0 | 0 | 0 | 1478 | 0 | 0 | 0 | 0 | 5 | 0 | 0 | 1 | 0 | 0 | 0 |
| **Influence** | 0 | 0 | 0 | 0 | 0 | 1 | 0 | 0 | 0 | 0 | 2 | 23 | 0 | 0 | 0 | 7 | 0 | 0 | 0 | 0 | 0 | 1 |
| **Interest** | 0 | 0 | 0 | 0 | 0 | 3 | 15 | 8 | 1 | 3 | 24 | 3 | 1201 | 13 | 10 | 62 | 28 | 0 | 1 | 0 | 0 | 6 |
| **Learning** | 0 | 0 | 0 | 0 | 0 | 0 | 8 | 1 | 0 | 1 | 3 | 1 | 0 | 4 | 3 | 7 | 0 | 0 | 0 | 0 | 0 | 1 |
| **No reason** | 0 | 1 | 0 | 0 | 0 | 0 | 0 | 2 | 1 | 0 | 6 | 2 | 0 | 0 | 19 | 6 | 1 | 0 | 0 | 0 | 0 | 4 |
| **Non-identifiable/Other** | 0 | 0 | 0 | 0 | 0 | 0 | 0 | 0 | 0 | 0 | 0 | 0 | 2 | 0 | 1 | 4 | 3 | 0 | 2 | 1 | 0 | 0 |
| **Other survey** | 0 | 0 | 0 | 0 | 0 | 0 | 1 | 0 | 0 | 0 | 0 | 0 | 0 | 0 | 0 | 2 | 1 | 0 | 0 | 0 | 0 | 0 |

| | | | | | | | | | | | | | | | | | | | | | |
|---|---|---|---|---|---|---|---|---|---|---|---|---|---|---|---|---|---|---|---|---|---|
| characteristics | | | | | | | | | | | | | | | | | | | | | |
| Professionalism | 0 | 0 | 0 | 0 | 0 | 0 | 1 | 1 | 1 | 0 | 0 | 0 | 0 | 0 | 0 | 1 | 3 | 23 | 0 | 0 | 0 | 0 |
| Recruiter | 0 | 0 | 0 | 2 | 0 | 0 | 0 | 0 | 0 | 0 | 1 | 0 | 0 | 0 | 0 | 2 | 0 | 0 | 10 | 1 | 0 | 0 |
| Recruitment | 0 | 0 | 0 | 10 | 1 | 0 | 0 | 2 | 5 | 0 | 29 | 5 | 0 | 1 | 17 | 17 | 4 | 0 | 6 | 67 | 3 | 0 |
| Routine | 0 | 0 | 0 | 2 | 0 | 0 | 0 | 0 | 0 | 0 | 5 | 0 | 0 | 1 | 7 | 1 | 1 | 0 | 0 | 0 | 66 | 0 |
| Tell opinion | 0 | 0 | 0 | 0 | 0 | 3 | 4 | 3 | 2 | 3 | 63 | 4 | 0 | 2 | 88 | 45 | 3 | 0 | 0 | 0 | 0 | 209 |
| NA | 0 | 0 | 0 | 3 | 0 | 0 | 8 | 1 | 1 | 3 | 76 | 2 | 7 | 0 | 95 | 12 | 3 | 1 | 0 | 2 | 5 | 5 |

*Table A10: Confusion matrix (actual vs. predicted categories) for Mistral under zero-shot prompting with descriptions.*

**Mistral** – few-shot

| R: Prediction / C: Reference | Anonymity | Brevity | Curiosity | Dutifulness | Fun | Help politicians | Help science | Help society | Help, not further specified | Importance in general | Incentive | Influence | Interest | Learning | No reason | Non-identifiable/Other | Other survey characteristics | Professionalism | Recruiter | Recruitment | Routine | Tell opinion |
|---|---|---|---|---|---|---|---|---|---|---|---|---|---|---|---|---|---|---|---|---|---|---|
| **Anonymity** | 27 | 0 | 0 | 0 | 0 | 0 | 0 | 0 | 0 | 0 | 7 | 0 | 0 | 0 | 1 | 3 | 0 | 0 | 0 | 0 | 0 | 0 |
| **Brevity** | 0 | 8 | 0 | 0 | 0 | 0 | 0 | 0 | 0 | 0 | 0 | 0 | 0 | 0 | 6 | 1 | 11 | 0 | 0 | 0 | 0 | 0 |
| **Curiosity** | 0 | 0 | 38 | 0 | 1 | 0 | 3 | 0 | 0 | 1 | 17 | 0 | 15 | 10 | 4 | 9 | 0 | 0 | 0 | 0 | 0 | 0 |
| **Dutifulness** | 0 | 0 | 0 | 33 | 0 | 0 | 0 | 1 | 0 | 0 | 0 | 0 | 0 | 0 | 0 | 2 | 0 | 0 | 0 | 0 | 0 | 0 |
| **Fun** | 0 | 0 | 0 | 0 | 375 | 0 | 0 | 0 | 0 | 0 | 0 | 0 | 2 | 0 | 0 | 0 | 5 | 0 | 0 | 0 | 0 | 0 |
| **Help politicians** | 0 | 0 | 0 | 0 | 0 | 12 | 0 | 0 | 0 | 0 | 1 | 1 | 1 | 0 | 0 | 2 | 0 | 0 | 0 | 0 | 0 | 0 |
| **Help science** | 0 | 0 | 0 | 0 | 0 | 3 | 74 | 3 | 7 | 12 | 2 | 1 | 2 | 0 | 1 | 49 | 1 | 1 | 0 | 0 | 0 | 0 |
| **Help society** | 0 | 0 | 0 | 0 | 0 | 4 | 0 | 94 | 10 | 6 | 2 | 12 | 2 | 1 | 35 | 44 | 1 | 0 | 0 | 0 | 0 | 0 |
| **Help, not further specified** | 0 | 0 | 0 | 2 | 0 | 0 | 4 | 6 | 21 | 0 | 18 | 1 | 0 | 0 | 12 | 5 | 0 | 0 | 0 | 0 | 0 | 0 |
| **Importance in general** | 0 | 0 | 0 | 0 | 0 | 0 | 1 | 6 | 0 | 42 | 9 | 2 | 0 | 1 | 0 | 12 | 3 | 0 | 0 | 0 | 0 | 0 |
| **Incentive** | 0 | 0 | 0 | 7 | 0 | 0 | 0 | 0 | 0 | 0 | 1572 | 0 | 0 | 0 | 18 | 7 | 0 | 0 | 1 | 0 | 0 | 0 |
| **Influence** | 0 | 0 | 0 | 1 | 0 | 3 | 0 | 0 | 0 | 0 | 1 | 31 | 0 | 0 | 0 | 10 | 0 | 0 | 0 | 0 | 0 | 2 |
| **Interest** | 0 | 0 | 0 | 0 | 0 | 0 | 0 | 1 | 0 | 0 | 28 | 0 | 1162 | 5 | 5 | 28 | 25 | 0 | 0 | 0 | 0 | 0 |
| **Learning** | 0 | 0 | 0 | 0 | 0 | 0 | 2 | 1 | 0 | 0 | 0 | 0 | 0 | 8 | 4 | 8 | 0 | 0 | 0 | 0 | 0 | 0 |
| **No reason** | 0 | 0 | 0 | 0 | 0 | 0 | 0 | 0 | 0 | 0 | 2 | 1 | 0 | 0 | 19 | 9 | 3 | 0 | 0 | 0 | 0 | 0 |
| **Non-identifiable/Other** | 0 | 0 | 0 | 0 | 0 | 0 | 0 | 0 | 0 | 0 | 28 | 0 | 2 | 3 | 50 | 4 | 1 | 0 | 0 | 0 | 0 | 0 |
| **Other survey characteristics** | 0 | 0 | 0 | 0 | 0 | 0 | 0 | 0 | 0 | 0 | 2 | 0 | 62 | 2 | 20 | 13 | 9 | 0 | 0 | 0 | 0 | 0 |

| | | | | | | | | | | | | | | | | | | | | | |
|---|---|---|---|---|---|---|---|---|---|---|---|---|---|---|---|---|---|---|---|---|---|
| Professionalism | 0 | 0 | 0 | 0 | 0 | 0 | 2 | 0 | 0 | 0 | 2 | 0 | 0 | 0 | 0 | 1 | 4 | 23 | 0 | 0 | 0 | 0 |
| Recruiter | 0 | 0 | 0 | 0 | 0 | 0 | 0 | 0 | 0 | 0 | 1 | 0 | 0 | 0 | 0 | 0 | 0 | 0 | 7 | 2 | 0 | 0 |
| Recruitment | 0 | 0 | 0 | 13 | 0 | 0 | 0 | 0 | 3 | 0 | 95 | 4 | 1 | 0 | 0 | 11 | 2 | 0 | 10 | 69 | 2 | 1 |
| Routine | 0 | 0 | 0 | 0 | 0 | 0 | 0 | 0 | 0 | 0 | 2 | 0 | 0 | 0 | 0 | 3 | 0 | 0 | 0 | 0 | 72 | 0 |
| Tell opinion | 0 | 0 | 0 | 0 | 0 | 0 | 2 | 0 | 1 | 2 | 3 | 3 | 1 | 4 | 147 | 32 | 0 | 0 | 0 | 0 | 0 | 227 |
| NA | 0 | 0 | 0 | 0 | 0 | 0 | 0 | 0 | 0 | 0 | 5 | 1 | 2 | 0 | 37 | 2 | 0 | 0 | 2 | 0 | 0 | 2 |

*Table A11: Confusion matrix (actual vs. predicted categories) for Mistral under few-shot prompting.*

6. *Fine-Tuning Performance*

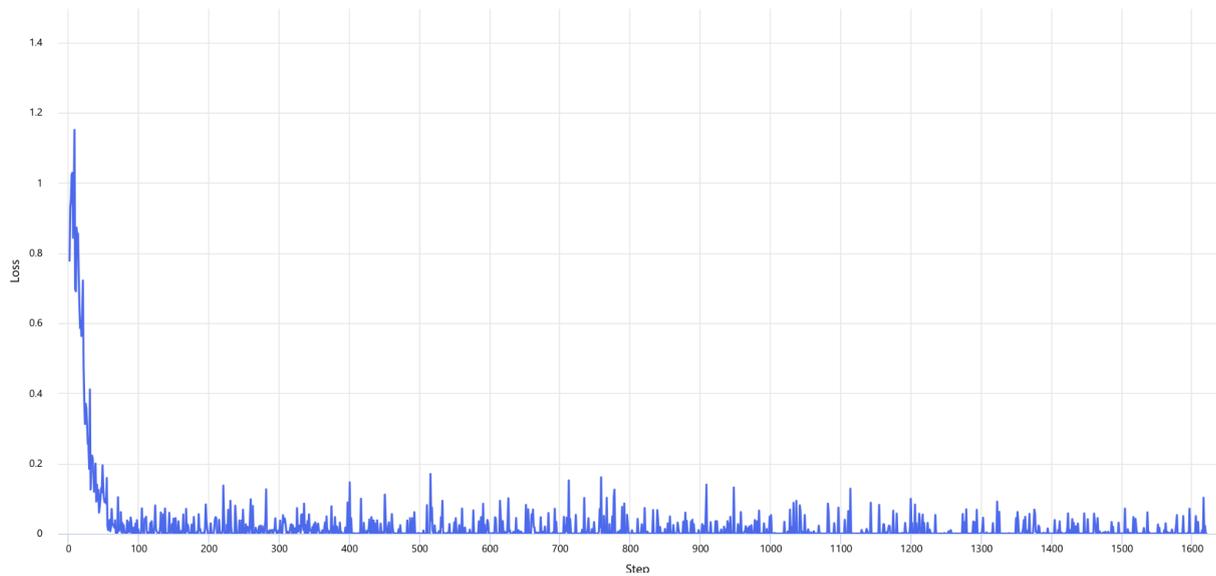

*Figure A7: Loss curve during fine-tuning.*
*Note: steps = n_examples/batch_size*n_epoch. Epochs describe the number of iterations through the data, and batch size the number of examples used in a single training pass.*

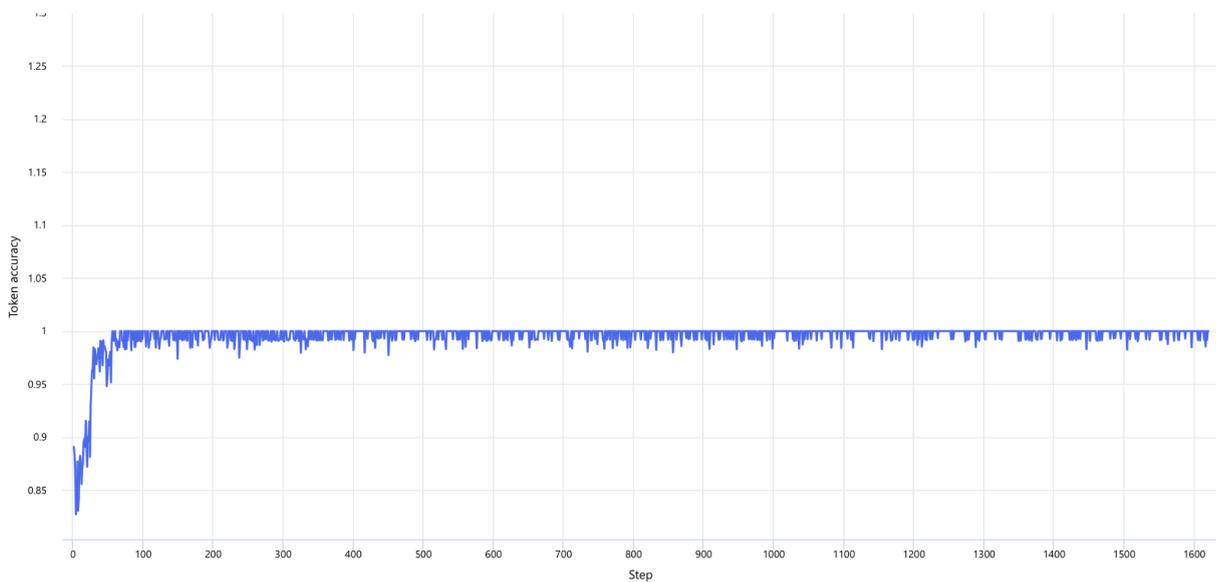

*Figure A8: Mean token accuracy achieved during fine-tuning.*